\documentclass[11pt]{article}
\usepackage{dirtytalk}

\usepackage{caption}
\usepackage[bottom, stable, hang,flushmargin]{footmisc}
\usepackage{pifont}
\usepackage{mathdots}

\usepackage{natbib}
\renewcommand{\thefootnote}{%
  \scalebox{1.3}{\textcircled{\raisebox{-0.4pt}{\kern0.2pt\tiny\arabic{footnote}}}}}

\makeatletter
\renewcommand\@makefntext[1]{%
  \noindent\makebox[1.8em][l]{\@makefnmark}#1}
\makeatother

\usepackage[utf8]{inputenc} 
\usepackage[T1]{fontenc}    
\usepackage{hyperref}       
\usepackage{url}            
\usepackage{booktabs}       
\usepackage{amsfonts}       
\usepackage{microtype}      
\usepackage{xcolor}         

\usepackage[preprint]{neurips_2026}
\usepackage[commenters={A,B,C,D,E,F,G,H,I,J}]{shortex}
\crefname{appendix}{Appendix}{Appendices}

\let\oldappendix\appendix
\renewcommand{\appendix}{%
  \oldappendix%
  \crefalias{section}{appendix}%
}
\author{%
  Maricela Best McKay\\
  University of British Columbia\\
  \texttt{maricela@math.ubc.ca} \\
  \And
  Nathan P. Lawrence \\
  University of California, Berkeley \\
  \texttt{nplawrence@berkeley.edu} \\
  \AND
  Brian Wetton \\
  University of British Columbia\\
  \texttt{wetton@math.ubc.ca} \\
  \And
  Bhushan Gopaluni \\
  University of British Columbia\\
  \texttt{bhushan.gopaluni@ubc.ca} \\
}

\title{Error whitening: Why Gauss-Newton outperforms Newton}
\date{Updated \today} 

\begin{document}

\maketitle
\begin{abstract}
	The Gauss-Newton matrix is widely viewed as a positive semidefinite approximation of the Hessian, yet mounting empirical evidence shows that Gauss-Newton descent outperforms Newton's method. We adopt a function space perspective to analyze this phenomenon. We show that the generalized Gauss-Newton (GGN) matrix projects the Newton direction in function space onto the model's tangent space, while a Jacobian-only variant obtained by applying the least squares Gauss-Newton matrix to non-least squares losses projects the function space loss gradient onto this same tangent space. Both projections eliminate distortions from the model's parameterization. Specifically, the evolution of the prediction-target mismatch depends on the model's parameterization through the matrix $JJ^\top$ where $J$ is the Jacobian of the model with respect to its parameters. The projections effectively replace $JJ^\top$ with the identity. We call this effect \emph{\textbf{error whitening}}. Once the parameterization is removed, the prediction-target mismatch evolves according to dynamics dictated by the structure of the loss and the projection produced by the optimizer. 
    Error whitening is a special property of Gauss-Newton descent that rigorously distinguishes it from Newton's method. 
    We empirically demonstrate that Gauss-Newton optimizers follow the theoretically predicted function space dynamics and outperforms Newton's method, Adam, and Muon across case studies spanning supervised learning, physics-informed deep learning, and approximate dynamic programming. 
\end{abstract} 
\section{Introduction}

Historically, the Gauss-Newton matrix was developed in the context of least squares problems (particularly nonlinear least squares) \cite{bjorck2024numerical, nocedal2006numerical, martens2020new}. The typical viewpoint is to motivate Gauss-Newton as a positive semidefinite approximation to the Hessian obtained by dropping second-order information from the network. This approximation is identical to the full Hessian at a minimum and is attractive in practice because it is less costly, as it requires only first-order derivatives of the neural network, which are already needed for backpropagation. 
Despite this close relationship between the Gauss-Newton matrix and the Hessian, it has been repeatedly observed that Gauss-Newton descent outperforms Newton's method \cite{martens2020new, jnini2025gauss, SNEGD}. This empirical gap is important to address in light of recent trends that argue for a ``second-order is better"  perspective. \cite{abreu2025potential} use full Gauss-Newton as a practical upper bound on iteration complexity for LLM pretraining and report large gains over SOAP \cite{vyas2024soap} and Muon \cite{large2024scalable}, both of which can themselves are often discussed as approximate second-order methods \cite{anil2020shampoo, morwani2024shampoo}, while \cite{wang2025gradient} appeal to a second-order perspective to explain various optimizers for physics-informed neural networks. 
Yet, to our knowledge, there is a surprising lack of insight in the literature to explain this phenomenon. A notable exception is \cite{chen2011hessian}, which shows that Gauss-Newton descent preserves important structural properties of the solution manifold that Newton's method does not.

We shed light on this phenomenon. In particular, we show that explicitly decoupling Gauss–Newton from the Hessian clarifies why Gauss–Newton–type updates can be more effective than Newton’s method in practice. This perspective motivates defining a Gauss–Newton–type matrix independent of the loss curvature: $$G_J = \frac{1}{d}\sum_{i=1}^d \lt(\nabla_\theta f_\theta(\textbf{x}_i)^{\top} \nabla_\theta f_\theta(\textbf{x}_i) \rt).$$ This matrix is the Gauss–Newton matrix that one gets from a least squares loss, but it can simply be applied to the gradients from a non-least squares loss.
We show that Gauss–Newton descent corresponds to a weighted projection of Newton's method in function space onto the model's tangent space, while $G_J$ projects the function space loss gradient onto the model's tangent space. In both cases, these projections align changes in $f_\theta$ more directly with the mismatch between predictions and the training targets. We term this behaviour \textbf{\textit{error whitening}} to evoke the similar phenomenon of gradient whitening \cite{fransstable,lu2025understanding,yang2008principal}.  

\paragraph{Main contributions.}
Our contributions are summarized as follows.
\begin{itemize}
    \item We introduce the notion of error whitening as a mechanism for understanding the fundamental difference between Gauss-Newton and Newton. We derive mismatch ODE dynamics with closed form solutions for specific losses that accurately predict empirical convergence. Our analysis provides a rigorous explanation for the superior performance of Gauss-Newton.
	\item Detailed case studies spanning supervised learning, physics-informed learning, approximate dynamic programming. These empirical results corroborate our theoretical claims and demonstrates broad improvements Gauss-Newton provides simply by changing the optimizer.
\end{itemize}

\paragraph{Problem setting.} Consider a dataset $\{(\textbf{x}_1, \textbf{y}_1), \dots , (\textbf{x}_d,\textbf{y}_d)\}$ drawn from inputs $\textbf{x}\in X \subset \reals^m$ and outputs $\textbf{y}\in Y \subset \reals^k$. We assume that the relationship between the inputs and outputs can be described by a mapping $f:X \mapsto Y$. In this setting, a neural network is a parametric function that approximates this mapping.  
Let $f_\theta: X \mapsto Y$ denote this neural network, where $\theta \in \reals^p$ denotes its weights. We seek the best approximation to $f$ over the family of functions $f_\theta$ via minimization of a loss function. In practice, rather than working with the expectation of the loss, because the distribution of the data is not known, it is common to minimize the empirical risk  \[ \label{e:empiricalrisk}
L(\theta) = \frac{1}{d} \sum_{i=1}^d \ell \lt(f_\theta(\textbf{x}_i),  \textbf{y}_i \rt).\]
The local loss function $\ell$ may be, for example, mean-squared error, cross-entropy, or a variety of other choices. It is almost always a convex function with respect to its argument $f_\theta$; however, when composed with the neural network, the loss landscape is nonconvex and can be challenging to optimize.

Minimizing the second-order Taylor expansion of $L$ about a small perturbation in the parameters $\Delta \theta$ yields Newton's method with update rule $ \theta_{k+1} = \theta_k - H^{-1}\nabla_\theta L(\theta)$,
where $H = \nabla_\theta^2 L$ is the Hessian of the loss.\footnotemark[1]\footnotetext[1]{When the matrix is singular, one should use the Moore-Penrose inverse denoted by $\dag$.}  Let $\ell_i := \ell\lt(f_\theta(\textbf{x}_i), \textbf{y}_i\rt)$. Using the chain rule, we can write the Hessian of \cref{e:empiricalrisk} as
\[ \label{e:Hessian}
\frac{1}{d}\sum_{i=1}^d \Big(
 \underbrace{\nabla^2_\theta f_\theta(\textbf{x}_i)}_{p \times p \times k} \underbrace{\nabla_{f_\theta} \ell_i}_{k\times 1} + 
\underbrace{\nabla_\theta f_\theta(\textbf{x}_i)^{\top}}_{p\times k} \underbrace{\nabla_{f_\theta}^2 \ell_i}_{k \times k}\underbrace{\nabla_\theta f_\theta(\textbf{x}_i)}_{k \times p}  
\Big).\footnotemark[2] \] \footnotetext[2]{Note that the tensor-vector contraction
$\nabla^2_\theta f_\theta(\textbf{x}_i)\nabla_{f_\theta} \ell_i=\sum_{j=1}^k \lt(\nabla^2_\theta f_\theta(\textbf{x}_i)\rt)_{[:,:,j]}(\nabla_{f_\theta}\ell_i)_{[j,:]}$ where $\lt(\nabla^2_\theta f_\theta(\textbf{x}_i)\rt)_{[:,:,j]}$, for example, is the Hessian of the $j$-th output component of $f_\theta$.}
In Newton's method, the Taylor expansion is truncated at order two, yielding a local quadratic model. However, this approximation is accurate only in a local neighbourhood of the parameters. For a sufficiently small neighbourhood, $f_\theta$ will be close to its linearization and the second-order term $\nabla^2_\theta f_\theta(\textbf{x}_i)\nabla_{f_\theta} \ell_i$ in \cref{e:Hessian} vanishes. This gives rise to the generalized Gauss-Newton (GGN) matrix,
\[ \label{GGN}
G = \frac{1}{d}\sum_{i=1}^d \nabla_\theta f_\theta(\textbf{x}_i)^{\top} \nabla_{f_\theta}^2 \ell_i \nabla_\theta f_\theta(\textbf{x}_i), 
\]  
which yields an analogous update rule $ \theta_{k+1} = \theta_k - G^{-1}\nabla_\theta L(\theta)$ to Newton's method.

Newton and Gauss-Newton be thought of as quadratic approximations of the loss landscape. Newton's method, however, unlike Gauss-Newton-type schemes, does not assume that the neural network can be treated as linear. What distinguishes Newton from Gauss-Newton is the term that incorporates second-order curvature information from $f_\theta$. The quadratic assumption relies on operating locally enough that the highly nonconvex loss landscape can be faithfully approximated by a quadratic in the local neighbourhood of a step. 
If the quadratic model is valid in a small enough neighbourhood of parameter space, then the model is actually approximately linear. Said slightly differently, $f_\theta$ is effectively linear along a small enough step. For such steps, this second-order term may be actively harmful. 
The rest of this paper builds on these observations and ultimately advocates for treating Gauss-Newton-type optimizers as divorced from the ethos surrounding classical Newton.

\section{Gauss-Newton in function space}
\label{sec:GNF}

In this section, we analyze Gauss-Newton descent from a function space perspective. We start by analyzing the GGN update, then show how replacing it with $G_J$ changes the story. This lays the groundwork for \cref{sec:error_whitening} where we will see how the function space behaviour of each model leads to \emph{error whitening}. 

We adopt a continuum viewpoint on optimization \cite{su2016differential}. Taking the step size $\eta \rightarrow 0$ in a gradient descent update gives the gradient flow $\frac{d \theta}{d \tau} = -\nabla_\theta L(\theta)$, an ordinary differential equation (ODE) governing the trajectory of parameters during training. One obtains an analogous ODE for Gauss-Newton descent: $\frac{d \theta}{d \tau} = -G^{\dag}\nabla_\theta L(\theta)$.
To understand model behaviour, we examine the evolution of the network $f_\theta$ along this trajectory. By the chain rule,
\[\frac{d f_\theta}{d\tau} = J\,\frac{d\theta}{d\tau},\label{eq:funcflow}\] where $J= \begin{bmatrix}
   \nabla_{\theta} f_\theta(x_1),
  \, \dots, \,
  \nabla_{\theta} f_\theta(x_d)
\end{bmatrix}^{\top} \in \mathbb{R}^{dk\times p}$ is the Jacobian of the stacked output vector $
\begin{bmatrix}
f_\theta(x_1)^\top,
\, \dots, \,
f_\theta(x_d)^\top
\end{bmatrix}^{\top}
\in \mathbb{R}^{dk}$. 

 \bprop \label{generalfunctionspaceflow} Suppose that $\ell$ is strictly convex and twice differentiable in $f_\theta$. Let $\nabla_{f_\theta} \ell(\mathbf{f}_{\theta})= \frac{1}{d} \begin{bmatrix}
   \nabla_{f_\theta} \ell_1,
  \, \dots, \,
  \nabla_{f_\theta}\ell_d
\end{bmatrix}^{\top} \in \reals^{dk}$ denote the vectorized function space gradient and $H_{\ell} = \nabla^2_{f_\theta}\ell(f_\theta) \in \mathbb{R}^{dk\times dk}$ be the block diagonal function space Hessian of the loss. Gauss–Newton descent corresponds to the Newton update direction in function space, restricted to directions reachable through parameter updates.
 Moreover, this direction is the unique minimizer of 
\[
\min_{v \in \mathrm{Im}(\nabla_\theta f_\theta)} \frac{1}{2} 
\left\| v + H_\ell^{-1} \nabla_{f_\theta} \ell(f_\theta) \right\|_{H_\ell}^2.
\]
\eprop
 \begin{proof}[Proof sketch] Let $A = H_\ell^{1/2} J$. Substituting the Gauss-Newton gradient flow into the right-hand side of \cref{eq:funcflow} we obtain
\[
\frac{d \mathbf{f}_\theta}{d\tau} =  -H_\ell^{-1/2}A A^\dag \lt(H_\ell^{-1/2}\nabla_{f_\theta}\ell(\mathbf{f}_{\theta})\rt).\\  \] Since $AA^\dagger$ is the Euclidean projector onto $\mathrm{Im}(A)$, conjugation by $H_\ell^{\pm 1/2}$ gives a projection in the $H_\ell$-weighted inner-product onto $\mathrm{Im}(J)$. The full proof is in \cref{app:proofs}.
 \end{proof}
 
\bcor \label{GNfuncgradient}
$G_J$ gives the closest approximation to $\nabla_{f_\theta} \ell(\mathbf{f}_{\theta})$ in the tangent space of $f_{\theta}$.
\bprf
For $G_J$, the function space Hessian $H_\ell$ is absent, and we simply get 
\[\frac{d \mathbf{f}_\theta}{d\tau}  = -J J^\dag \nabla_{f_\theta}\ell(\mathbf{f}_{\theta}).\label{e:GJfuncspace}
 \]Here, $JJ^\dag$ is the orthogonal projector that takes $\nabla_{f_\theta}\ell(\mathbf{f}_{\theta})$ to the column space of $J$. Note that, unlike the GGN, this projection is in the standard Euclidean norm.
 \eprf	
\ecor
The behaviour of these projections plays an important role in the next section, where we define the central topic of this paper: error whitening.

\section{Error whitening}
\label{sec:error_whitening}

Let $r = \psi(\mathbf{f}_\theta, \mathbf{y})$ where $\psi$ is a function that measures a notion of mismatch. We can rewrite the loss as a function of the mismatch  $\ell = \hat{\ell}(r)$. Then, by the chain rule
\[\nabla_{f_\theta} \ell = \nabla_{f_\theta} \psi\rbra*{\mathbf{f}_\theta, \textbf{y}}^{\top} \nabla_r \hat{\ell}(r).\] 
In words: this is the sensitivity of the mismatch to the model output, multiplied by the sensitivity of the loss to the mismatch. The vector $\nabla_r \hat{\ell}(r)$ measures how strongly the mismatch affects the loss, and the Jacobian $\nabla_{f_\theta} \psi\rbra*{\textbf{f}_\theta, \textbf{y}}$ maps this information back to the space of model outputs.

To be effective, an optimization scheme must drive the mismatch to zero. However, we can't directly control what happens to the mismatch. What we directly change are the parameters $\theta$, and these are updated according to a loss function. Between the parameters and the mismatch lies a sequence of transformations:
\[
\theta \xrightarrow{J} f_\theta \xrightarrow{\nabla_{f_\theta}\psi} r \xrightarrow{\nabla_r\hat{\ell}} \ell.
\]
These appear in the structure of the gradient $\nabla_\theta \ell = J^{\top} \nabla_{f_\theta}\psi^{\top} \nabla_r \hat{\ell}$. Consider how the mismatch evolves along the training path: 
\[
\frac{dr}{d\tau} = \nabla_{f_\theta}\psi(\mathbf{f}_\theta, \mathbf{y}) \frac{df_\theta}{d\tau}.
\label{e:mismatch_evolution}
\]

The right-hand side of \cref{e:mismatch_evolution} is determined by the function space dynamics in \cref{eq:funcflow} of the last section:
\begin{align}
\text{Gradient descent:} \quad \frac{dr}{d\tau} &= -\nabla_{f_\theta}\psi \, JJ^\top \, \nabla_{f_\theta}\psi^\top \, \nabla_r\hat{\ell}(r) \label{e:mismatch_gd} \\
G_J: \quad \frac{dr}{d\tau} &= -\nabla_{f_\theta}\psi \, JJ^\dag \, \nabla_{f_\theta}\psi^\top \, \nabla_r\hat{\ell}(r) \label{e:mismatch_gj} \\
\text{GGN:} \quad \frac{dr}{d\tau} &= -\nabla_{f_\theta}\psi \, H_\ell^{-1/2}AA^\dag H_\ell^{-1/2} \, \nabla_{f_\theta}\psi^\top \, \nabla_r\hat{\ell}(r). \label{e:mismatch_ggn}
\end{align}

\begin{definition}[Error whitening]
An optimization method is \textbf{\emph{error whitening}} if it replaces $JJ^\top$ in \cref{e:mismatch_gd} with an operator that cancels $J$'s singular values. 
\end{definition}

The term \emph{error whitening} is inspired by gradient whitening, where the parameter space gradient $g$ is replaced by a $\tilde{g}$ such that $\mathbb{E}[\tilde{g}\tilde{g}^T] = I$. Analogously, in our definition, the nonzero eigenvalues of the Jacobian Gram matrix $JJ^T$ are removed. Unlike gradient whitening, which acts in parameter space, error whitening acts in function space.

For $G_J$ in \cref{e:mismatch_gj}, the analogy with gradient whitening is clearer: all of $JJ^\top$'s nonzero eigenvalues become $1$, so $JJ^\top$ is replaced by the identity on $\mathrm{Im}(J)$ (and by the identity on all of $\mathbb{R}^{dk}$ when $J$ has full row rank). In \cref{e:mismatch_ggn}, GGN also cancels $J$'s singular values, but this can be read as a two-step composition: first $JJ^\top$ is replaced by the identity on $\mathrm{Im}(J)$, and then this identity is rescaled by $H_\ell^{-1}$, producing a function-space Newton update on $\mathrm{Im}(J)$. The composition is exact when $J$ has full row rank. This situation occurs when the network is overparameterized, and the training data are linearly independent. For both optimizers, removing $J$'s singular values removes one level of distortion in the sequence of transformations between the parameters and the mismatch. 
We consider what this looks like for several concrete examples. In the examples below, we derive the mismatch dynamics under the idealized case where $J$ has full row rank, and the projections can be ignored.
%


\paragraph{Squared and $q$-power losses.}
For $\ell_i = \tfrac{1}{q}(f_\theta(\mathbf{x}_i)-\mathbf{y}_i)^q$ and $q\ge 2,$
we have $\psi(f_\theta(\mathbf{x}_i),\mathbf{y}_i)=f_\theta(\mathbf{x}_i)-\mathbf{y}_i$, and $\hat\ell(r)=\tfrac{1}{q}r^q$. Because $\nabla_{f_\theta} \psi = \mathbf{I}$ we get

\begin{align}
G_J: \quad \frac{dr}{d\tau} &= - r^{q-1} &
\text{GGN:} \quad \frac{dr}{d\tau} &= - \frac{1}{q-1}r. 
\end{align}

\begin{figure}[tbh]
\begin{center}
  \includegraphics[width=0.7\columnwidth]{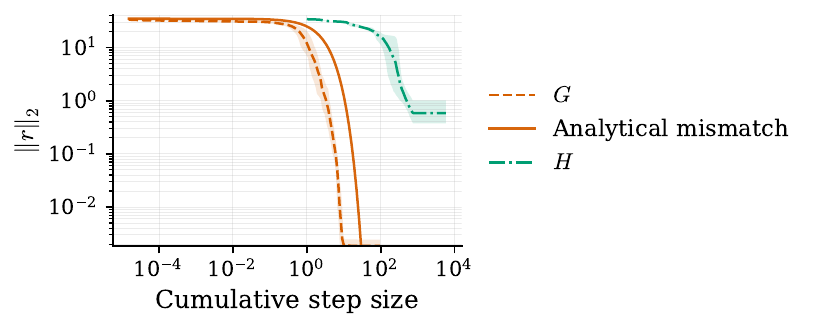}
  \end{center} 
  \caption{Evolution of the residual for a loss with $q=4$. This is an empirical realization of the GGN mismatch dynamics. See \cref{funcregl4} for details.}
  \label{fig:l4residualdecay}
\end{figure}

For a least squares problem where $q=2$, the GGN matrix and $G_J$ coincide with $\quad \frac{dr}{d\tau} = - r$. The solution to this ODE is $r(\tau) = e^{-\tau} r(0)$, which attains exponential decay. \Cref{fig:l4residualdecay} shows that the idealized analytical mismatch dynamics predict the empirical convergence of the mismatch and that Newton's method ($H$) has distinct mismatch dynamics.

For $q>2$, we still have an ODE with a right-hand side proportional to $r$ and the rate of decay will still be exponential. Under $G_J$, the mismatch ODE has solution 
\[\text{diag}\lt(\lt(1+ (q-2) r_i(0)^{(q-2)} \tau\rt)^{-1/(q-2)}\rt)r(0).\]
Different components of the residual decay at different rates, but for all of these, the rate is algebraic.

\paragraph{Cross-entropy.}
Let $\mathbf{p} = \mathrm{softmax}(f_\theta(\mathbf{x}))$, then for classification with one-hot labels $\mathbf{y}$ and logits $f_\theta(\mathbf{x})$, mismatch is given by $
\psi(f_\theta(\mathbf{x}),\mathbf{y})=\mathbf{p}-\mathbf{y}.$ Since $\psi$ is no longer a linear function, $\nabla_{f_\theta} \psi$ is no longer the identity. Instead it equals $ F:= $ diag$(\textbf{p})-\textbf{p}\textbf{p}^\top \neq \textbf{I}$, the well-known Fisher information matrix. For cross-entropy, $\ell$ is the negative log-likelihood, and $H_\ell= F$ as well. The idealized mismatch ODEs are:
\begin{align}
G_J: \quad \frac{dr}{d\tau} &= -Fr &
\text{GGN:} \quad \frac{dr}{d\tau} &= -r.
\end{align}
The solution for the GGN is again $r(\tau) = r(0)e^{-\tau}$ so the decay is uniform and exponential for all components of the residual. For $G_J$, $F$ is not independent of $r$, so writing $\textbf{p} = r+\textbf{y}$ we have $F = $ diag$\lt((r+\textbf{y}) - \lt((r+\textbf{y})(r+\textbf{y})^\top\rt)\rt)$. $G_J$'s mismatch ODE is a coupled nonlinear system. $F$ determines the coupling and the rate, but $F$ is a function of $r$, and its varying eigenstructure changes the rate of decay throughout training. As $\textbf{p} \rightarrow \textbf{y}$, the eigenvalues of $F$ vanish and the rate of decay of $r$ slows.

\paragraph{Function space alignment.}
All of the idealized mismatch rates depend on function space dynamics.
For squared, and other $q$-power losses, we have $\nabla_{f_\theta}\psi =
\mathbf{I}$, so the function space update and the mismatch update are one
and the same. When the function space update is proportional to $-r$, exponential decay follows. For cross-entropy, $G_J$'s function space update is proportional to $r$, but the decay rate is governed by $-Fr$. Under GGN, to realize the ideal exponential decay of the mismatch, the function space should align with $-F^{-1}r$. For cross-entropy, $\ell$ is not strictly convex, so $F$ is positive semidefinite rather than strictly positive definite, and the idealized (nonprojected) function space update for GGN aligns with $-F^\dag r$. 

\section{Scalable Gauss-Newton through sketching}
\label{sec:sketching}

Newton's method and Gauss-Newton are not widely used in deep learning because both require working with a matrix with $|\theta|^2$ entries. Many approaches circumvent this, including limited-memory quasi-Newton methods like L-BFGS \cite{liu1989limited}, layerwise Kronecker-factored approximations like K-FAC \cite{martens2015optimizing}, and iterative solvers that never form the full matrix. K-FAC assumes that the underlying matrix is block diagonal and further enforces a fixed Kronecker form on each layer's block. L-BFGS can only approximate the full Hessian and it retains curvature only along recently traversed directions. To make the optimization tractable we use the sketching methodology of \cite{SNEGD} applied to $G$, $G_J$ and to the full Hessian of the loss. Sketching approximates all eigenvalues above a tolerance, while iterative solvers work top down and only as far as an iteration budget permits. Because each of these methods is a quadratic approximation to a non-quadratic loss valid only locally, and the sketching methodology may introduce further error, we employ a grid-line search to select the step size at each iteration. 

Another important artifact of the sketching framework is that it produces a positive semi-definite (PSD) approximation of the matrix being sketched. At each training iteration approximate eigenvalues below a positive tolerance are set to zero. The Hessian is not necessarily PSD and so is not guaranteed to be a descent direction and may get stuck at saddle-points. It is standard practice to modify the Hessian to be PSD to avoid this issue \cite{nocedal2006numerical}, and clipping negative eigenvalues to zero is a well-known way to achieve this \cite{dauphin2014identifying, fletcher2013practical}. Note that, the PSD portion of the Hessian includes more than just $G$ \cite{martens2020new}. More details on the sketching methodology are included in \cref{app:sketch}. 

\section{Case studies}
\label{sec:examples}

 \Cref{generalfunctionspaceflow} shows that the GGN projects a Newton direction in function space onto the model's tangent space in the $H_\ell$ norm, while \cref{GNfuncgradient} shows that $G_J$ projects the loss gradient onto the tangent space in the standard Euclidean norm. In \cref{sec:error_whitening}, we see that the behaviour of these projections depends on the structure of the loss. In this section, we address the following questions: Does the error-whitening behaviour predicted by the function space perspective actually manifest during training, and does it explain the performance gap between Gauss–Newton methods and Newton's method?
We answer in the affirmative on case studies covering supervised learning, physics-informed learning, and reinforcement learning-type problems (approximate dynamic programming). 

For each task, we train a network using five different optimization schemes,
Gauss-Newton descent under $G$ and $G_J$, Newton's method, Adam \cite{ADAM}, and Muon \cite{large2024scalable}.
The experiments are kept as close as possible to enable a fair comparison across optimizers. 
All hyperparameters, except those related to the optimization, are kept the same.
Hyperparameters, additional experiments, and other supplementary material are included in \cref{app:losses,app:supervised,app:mnist,app:rl}.
All experiments are implemented in JAX \cite{jax2018github}.
Code is openly available.\footnote{\url{https://github.com/MaricelaM/error-whitening}}

\subsection{Regression} \label{funcregl4}

The first case study is a supervised regression task with a scalar-valued target function $g: \mathbb{R}^2\to\mathbb{R}$. We train a neural network $f_\theta$ on samples drawn from the unit square to approximate the function $g(x,y) = \sin(2 \pi x) \sin(2\pi y) + \sin(7\pi x)\sin(7\pi y)$. To properly compare a GGN and $G_J$ update, we need a nontrivial function space loss curvature, i.e., $H_\ell \neq \mathbf{I}$. To that end, we compare two loss functions: quartic loss, which should produce a marked difference in function space behaviour according to the analysis in \cref{sec:error_whitening}, and the log-cosh loss, which should not. For supervised regression, the mismatch is the residual $\textbf{f}_\theta -\textbf{y}$, so $\nabla_{f_\theta} \psi = \textbf{I}$ and the function space update has the same behaviour as the mismatch decay. 
In \cref{app:mnist}, we also study the MNIST \cite{lecun2002gradient} benchmark to demonstrate the case where the mismatch dynamics and the function space dynamics are not the same. 

\paragraph{Function space snapshots.} \Cref{fig:funcspace_l4} shows the function space update directions for all optimizers computed using parameters along a Muon training path at two loss levels. To compare the directions meaningfully, we need to use the same set of parameters for all optimizers; each optimizer induces a different path in parameter space, leading to a different mismatch function. To address this, we chose snapshots along Muon's optimization path because they provide a trajectory independent of any of the matrices whose function space behaviour we want to contrast. The snapshot tells us: starting at this point in parameter space, where would each optimizer's update direction point in function space?  

\begin{figure}[tbh]
\begin{center}
  \includegraphics[width=\columnwidth]{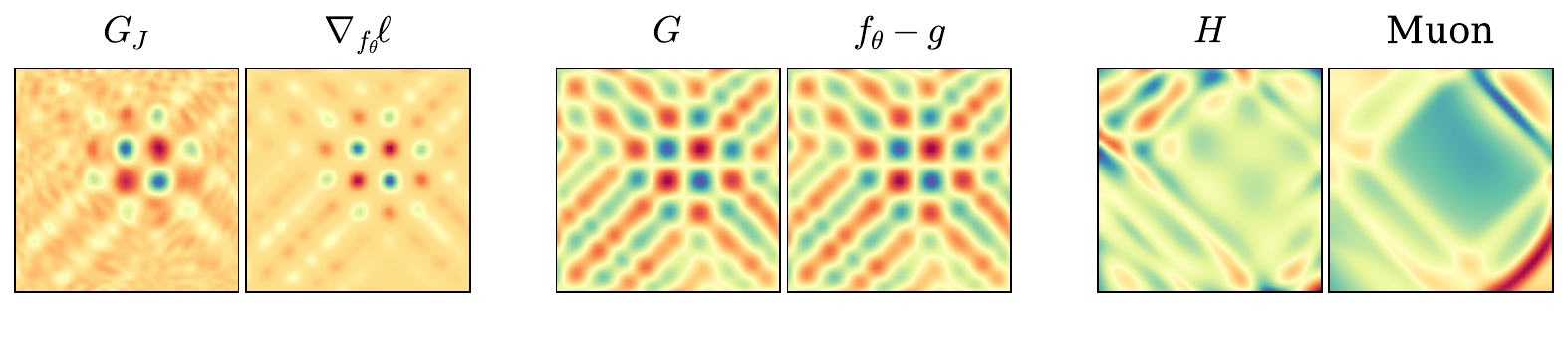}
  \end{center} 
  \caption{Function space update directions for each optimizer, visualized as heat maps over the input domain, computed along a Muon training path at loss $\approx 1E-02$, compared against the mismatch and function space loss gradient. All directions are normalized.}
  \label{fig:funcspace_l4}
\end{figure}

\Cref{fig:funcspace_l4}'s panels show that $G_J$ visually aligns with $\nabla_{f_\theta} \ell$ and $G$ visually aligns with the mismatch $f_\theta -g$. Newton's method and Muon, on the other hand, don't line up with these directions in function space. 
This shows that the second-order term $\nabla^2_\theta f_\theta$ dropped by the GGN is nonnegligible and is steering $H$ away from the mismatch. \Cref{fig:l4residualdecay} in \cref{sec:error_whitening} also  demonstrates this. It shows the evolution of the residuals for $G$ and $H$ in this case study as a function of the cumulative step size. The step sizes selected via line-search at every iteration are cumulatively summed, i.e. $\tau = \sum_k \eta_k$ to enable comparison of the empirical residual convergence to the analytical mismatch ODE solution for $G$. While $G$'s residual trajectory matches that of the analytical solution, $H$ is doing something distinct.

For quartic loss the function space gradient, $\nabla_{f_\theta}\ell = \lt(f_\theta - g\rt)^3$, so the gradient amplifies large errors. We can see this behaviour in the panels of \cref{fig:funcspace_l4}, where the mismatch has more uniform features for both loss levels, while the gradient has sharper features.

\begin{figure}[tbh]
\begin{center}
  \includegraphics[width=\columnwidth]{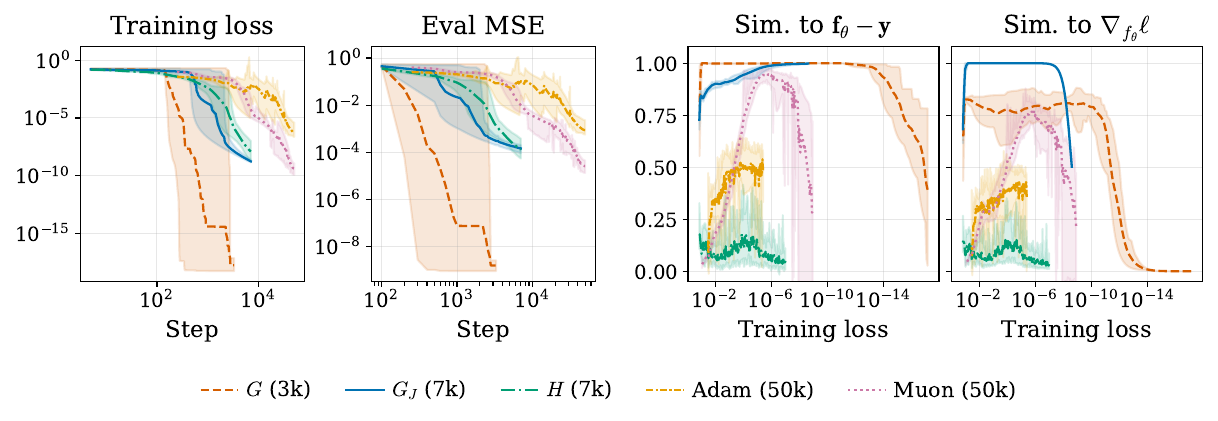}
  \end{center}
  \caption{Training loss, evaluation MSE, and cosine similarity to the mismatch and function-space loss gradient, for mean quartic loss. Cosine similarities are computed along each optimizer's respective training path. Solid lines show the mean across 10 random seeds; shaded regions span the min to max.} \label{fig:main_l4}
\end{figure}

\paragraph{Training alignment.} We now turn our attention to the entire training trajectory for each optimizer. \Cref{fig:main_l4} shows the cosine similarities between each optimizer's function space update and the mismatch and function space loss gradient as a function of the training loss. $G$ maintains high alignment with the mismatch until convergence when the residual is very small, and there is nothing to align with. $G_J$, on the other hand, aligns nearly perfectly with $\nabla_{f_\theta} \ell$ until the end of training. Muon, notably, reaches fairly high alignment with the mismatch along its own training path, which we also saw in the lower loss snapshot in
(\cref{tab:funcspace-cosine}). However, this alignment emerges only as Muon approaches convergence, and Muon does not reach the low error levels attained by $G$, which is naturally aligned with the mismatch for this loss at any point in parameter space. We will revisit this observation when discussing convergence below. 
Adam and Newton both have low alignment with the mismatch and the function space gradient. At first glance, the snapshot for loss 1E-05 and \cref{fig:main_l4} seem to be in contradiction. If a loss level of 1E-05 was enough to make the second-order term in the Hessian small for the snapshot, why do we not observe the same behaviour at a similar loss along the Newton training trajectory, and consequently a high alignment with the mismatch? 

To investigate this further, we compute the cosine similarity between the $G$ direction and the Newton direction along the trajectories induced by each optimizer for the same random seed. If the second-order term $\nabla^2_\theta f_\theta \nabla_{f_\theta} \ell$ is negligible, this cosine similarity should be close to one. At the Muon snapshot, the cosine similarities are 0.31 and 0.90 at loss levels 1E-02 and 1E-05, respectively. However, along $H$'s own training trajectory, the cosine similarity between the GGN and the Hessian is only 0.24 at a loss level of 1E-02 and 0.65 at a loss level of 1E-05, the lowest among all optimizers at similar loss levels. By contrast, along $G_J$'s path, the agreement is 0.79 and 0.97 at similar loss levels. Newton's method seems to follow a trajectory in parameter space where $\nabla^2_\theta f_\theta \nabla_{f_\theta} \ell$ remains more active. 

\paragraph{Convergence.}
\Cref{fig:main_l4} shows that aligning with the mismatch matters for convergence. $G$, which aligns with the residual, converges to a significantly lower loss and mean squared error than the other optimizers. Both the loss and the error are orders of magnitude lower. The gap between $G$ and $G_J$ is stark for this loss. $G_J$ aligns with the function space gradient, which amplifies large residuals and gets stuck as a result. $G_J$ and $H$ perform similarly. To ensure a fair comparison, Muon and Adam, which tend to require more iterations to exhibit convergence, are given more iterations than the sketched optimizers. Even though Muon aligns with the mismatch, especially as it converges, it isn't able to achieve the low loss and error that $G$ achieves.
\subsection{A physics-informed neural network}
\label{sec:pinn}

Physics-informed neural networks (PINNs) are a framework for approximating solutions to differential equations (DEs) using deep learning \cite{raissiPhysicsinformedNeuralNetworks2019}. They have been applied across a wide range of scientific and engineering problems, but PINNs are notoriously difficult to train, motivating active interest in improving their optimization \cite{rohrhofer2022role, wong2022learning, krishnapriyan2021characterizing, wangUnderstandingMitigatingGradient2021, wangWhenWhyPINNs2022b, daw2023mitigating}. To mitigate these difficulties, Gauss-Newton methods have been applied to PINNs. \citet{mullerAchievingHighAccuracy} introduced energy natural gradient descent (ENGD) which is equivalent to a Gauss-Newton update in the variational setting and when the underlying DE is linear. \cite{jnini2025gauss} directly apply Gauss-Newton to a non-linear PDE and observe that ENGD does worse. The Hessian of the loss for a PINN contains terms that ENGD drops. In this case study, we demonstrate that Gauss-Newton is effective for tackling a challenging PINN benchmark problem prevalent across the PINN literature, the Allen-Cahn equation, while Newton fails to train for this problem.

For this problem, it is common to employ a variety of techniques in order to get any traction. Among these are, hard enforcement of boundary conditions, Fourier embedding, loss re-weighing, causal training, and the use of specialized architectures \cite{wang2024piratenets,  guilhoto2024deep, wang2025gradient, wang2022respecting, DONG2021110242}. By contrast, we use a relatively small feedforward neural network. The setup is identical to that of the seminal PINN paper \cite{raissiPhysicsinformedNeuralNetworks2019} except here we randomly resample points according to the R3 strategy from \cite{daw2023mitigating}. The goal is to isolate and highlight the role of the optimizer rather than other algorithmic designs. As is most common in PINNs we employ $q$-power loss with $q=2$. In this setting, $\nabla_{f_\theta}\psi =
\mathbf{I}$. 

For this case study, Newton's method fails to train, even when given twice as many iterations as Gauss-Newton (see \cref{tab:PINNerr} in \cref{app:PINN}). Simply using Gauss-Newton with resampling achieves the same order of magnitude error as the more elaborate and layered techniques in \cite{wang2024piratenets, guilhoto2024deep} and two orders of magnitude lower error than just R3 resampling on this problem \cite{daw2023mitigating}.
\begin{figure}[tbh]
\begin{center}
  \includegraphics[width=0.8\columnwidth]{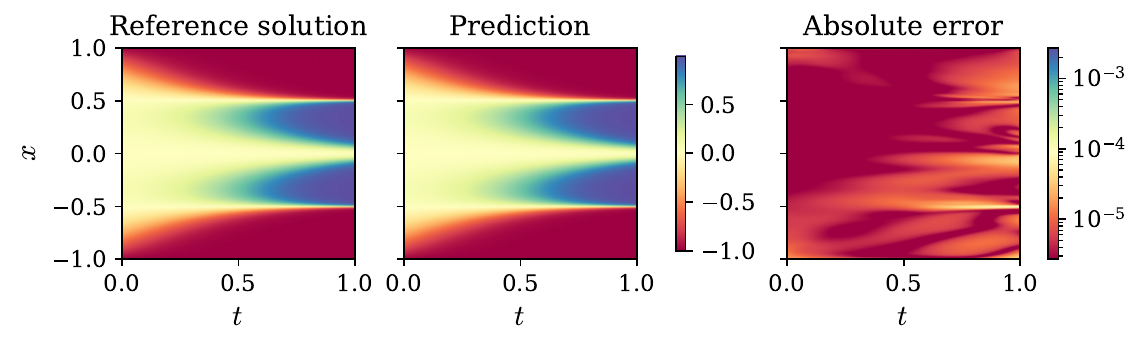}
  \end{center}
\caption{The reference Allen-Cahn solution, the PINN approximation achieved when training with $G$ and the absolute error between the two on a log scale in order to make differences visible.} 
\label{fig:AC}
\end{figure}

\subsection{Reinforcement learning}
\label{sec:rl}

This section demonstrates the effectiveness of Gauss-Newton optimization for value function estimation.
The value function is a core component of dynamic programming \cite{bertsekas2012dynamic}, and by extension, approximate dynamic programming \cite{powell2011ApproximateDynamic} and reinforcement learning \cite{sutton2018ReinforcementLearning}.
It predicts the future performance of a policy, making it a useful object for designing learning schemes that iteratively improve decision-making performance over time.
However, estimating the value function can be highly unstable, especially in data-driven settings such as reinforcement learning (RL).
Key sources of instability are known as the ``deadly triad'' \cite{sutton2018ReinforcementLearning}.
Namely, the use of (1) function approximators (neural networks) in tandem with (2) bootstrapped value estimates of the (3) (off-policy) optimal value function is a recipe for unstable learning.
This example explicitly demonstrates the training stability of Gauss-Newton optimization on its own without any stabilization techniques from the RL literature
\cite{hessel2017RainbowCombining, shengyi2022the37implementation, hasselt2010double, fujimoto2018AddressingFunction, wang2016dueling, lillicrap2015ContinuousControl, sutton1988LearningPredict, mnih2016AsynchronousMethods, schulman2017ProximalPolicy, schulman2015TrustRegion, lin1992reinforcement, mnih2013PlayingAtari, schaul2015prioritized, fujimoto2018AddressingFunction, haarnoja2018Softactorcritic}.
Further details regarding RL background and the problem setting are given in \cref{app:rl}.

\begin{figure}[tbh]
\begin{center}
  \includegraphics[width=\columnwidth]{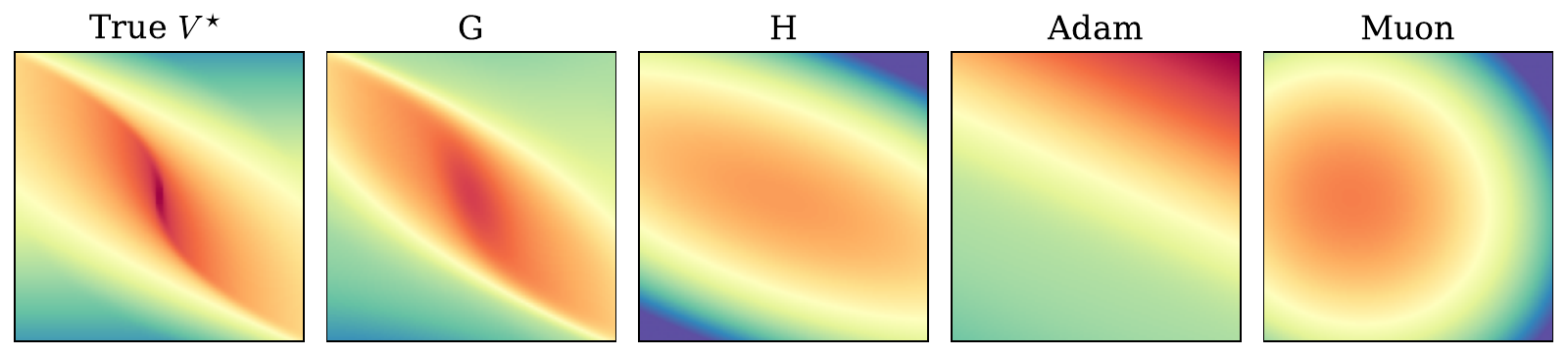}
  \end{center}
\caption{The true optimal value function compared against approximations obtained by each optimizer used in vanilla neural fitted value iteration. \Cref{fig:value_policy} further shows how the learned value functions align with the ground truth optimal actions.} 
\label{fig:value_function}
\end{figure}

We consider an offline neural fitted value iteration pipeline for the minimum-time problem applied to a double integrator environment.
This optimal control problem has an analytical bang-bang solution, making it a canonical example for comparing solution methods \cite{underactuated}.
The training data are uniformly sampled states around the origin (the goal state).
Unlike the previous examples, there is no static training target.
Instead, the neural network $V_\phi$ is used to compute its own targets:
\begin{equation}
	y(s) = \min_{a \in \{-1, 1\}} \left\{ \left(1 + V_\phi(f(s, a))\right)\1_{\left\{\norm{s} > \epsilon\right\}} (s) \right\},
\label{eq:value_target}
\end{equation}
where $f$ is the state transition function.
We apply vanilla neural fitted value iteration: targets in \cref{eq:value_target} are computed (and frozen), then $\phi$ is updated with one step of a given optimizer applied to the MSE. With these new parameters, targets are recomputed, and the process repeats.
\Cref{fig:value_function} shows the learned $V_\phi$ for each optimizer.
In essence, the improved alignment of Gauss-Newton translates to improved learning stability: Adam and Muon fail to capture the basic structure of the optimal value function; Newton's method captures a coarse profile of the value function; Gauss-Newton is the only optimizer to obtain the signature `S' shape of $V^\star$.   
\Cref{fig:value_policy} in \cref{app:rl} is a companion figure showing the estimation error and policy agreement among these value functions.
\Cref{app:rl} gives another case study, the inverted pendulum, for which there is not an analytical solution, further motivating the use of Gauss-Newton-type optimizers for RL problems.

\section{Conclusion and limitations}
We advocate for viewing Gauss-Newton methods as distinct from Newton's method rather than as less expensive approximations to Newton. The function space perspective explains the empirical performance gap between the two, and our case studies confirm that the predicted behaviour holds beyond the instantaneous-step limit.
 
 Our function space analysis describes instantaneous dynamics, i.e. the direction of a single infinitesimal step. This viewpoint is natural for Gauss-Newton methods, which rely on the linearization of $f_\theta$, and the empirical evidence confirms that $G_J$ and $G$ closely follow their instantaneous dynamics. For momentum-based optimizers like Adam and Muon, the one-step direction does not tell the full story: these methods reduce the loss and mismatch even when their instantaneous updates do not align with the ideal direction. Understanding mismatch dynamics beyond the one-step limit is left to future work. Sketching makes Gauss-Newton tractable for our examples, but problems with richer inherent structure will require higher ranks, and the approach does not scale to the largest network sizes. Our PINN and reinforcement learning case studies are proofs of concept that gesture toward a new starting point centered on robust optimization over heuristic stabilization techniques.

\bibliography{main}

\newpage
\appendix

\section{Identities and useful calculations}
\label{app:calcs}

\blem 
\label{lemmaP}
For any matrix $M \in \reals^{m\times n}$ with rank $r$, $(M^{\top}M)^\dag M^{\top} = M^\dag$. This is a standard well known result included here simply for completeness.
\elem
\bprf Consider the singular value decomposition of $M= U \Sigma V^{\top}$. Substituting the singular value decomposition into the left-hand side of the identity, we have
\[
(M^{\top}M)^\dag M^{\top} &= \rbra*{V\Sigma U^{\top} U \Sigma V^{\top}}^\dag V \Sigma U^{\top} \\
& = \rbra*{V \Sigma^{\top} \Sigma V^{\top}}^\dag V \Sigma^{\top} U^{\top}\\
& = V \rbra*{\Sigma^{\top}\Sigma}^\dag V^{\top} V \Sigma^{\top} U^{\top} \\
&= V
\begin{bmatrix}
\underbrace{
\begin{pmatrix}
  \sigma_1^{-2} &0   & \cdots & 0 \\
  0       & \sigma_2^{-2}& \ddots & \vdots \\
  \vdots         & \ddots & \ddots &  \\
  0              & \dots & 0& \sigma_r^{-2}  \\
\end{pmatrix}
}_{r\times r \text{ nonzero block}}
& \underbrace{\textbf{0}}_{r\times (n-r)} \\
\underbrace{\textbf{0}}_{(n-r)\times r} & \underbrace{\textbf{0}}_{(n-r)\times (n-r)}
\end{bmatrix}
\begin{bmatrix}
\underbrace{
\begin{pmatrix}
  \sigma_1 &0   & \cdots & 0 \\
  0       & \sigma_2& \ddots & \vdots \\
  \vdots         & \ddots & \ddots &  \\
  0              & \dots & 0& \sigma_r  \\
\end{pmatrix}
}_{r\times r \text{ nonzero block}}
& \underbrace{\textbf{0}}_{r\times (m-r)} \\
\underbrace{\textbf{0}}_{(n-r)\times r} & \underbrace{\textbf{0}}_{(n-r)\times (m-r)}
\end{bmatrix}U^{\top}\\
& = V
\begin{bmatrix}
\underbrace{
\begin{pmatrix}
  1/\sigma_1 &0   & \cdots & 0 \\
  0       & 1/\sigma_2& \ddots & \vdots \\
  \vdots         & \ddots & \ddots &  \\
  0              & \dots & 0& 1/\sigma_r  \\
\end{pmatrix}
}_{r\times r \text{ nonzero block}}
& \underbrace{\textbf{0}}_{r\times (m-r)} \\
\underbrace{\textbf{0}}_{(n-r)\times r} & \underbrace{\textbf{0}}_{(n-r)\times (m-r)}
\end{bmatrix}U^{\top} \\
& = V \Sigma^{\dag} U^{\top} \\
& = M^\dag .
\]
\eprf

\bcor \label{RightPseudoInverseID}
\[
	MM^\dag &= U\Sigma V^{\top} V\Sigma^\dag U^{\top} \\
	& = U \Sigma \Sigma^\dag U^{\top} \\
	& = U 
\begin{bmatrix}
\underbrace{
\begin{pmatrix}
  \sigma_1 &0   & \cdots & 0 \\
  0       & \sigma_2& \ddots & \vdots \\
  \vdots         & \ddots & \ddots &  \\
  0              & \dots & 0& \sigma_r  \\
\end{pmatrix}
}_{r\times r \text{ nonzero block}}
& \underbrace{\textbf{0}}_{r\times (n-r)} \\
\underbrace{\textbf{0}}_{(m-r)\times r} & \underbrace{\textbf{0}}_{(m-r)\times (n-r)}
\end{bmatrix}
\begin{bmatrix}
\underbrace{
\begin{pmatrix}
  1/\sigma_1 &0   & \cdots & 0 \\
  0       & 1/\sigma_2& \ddots & \vdots \\
  \vdots         & \ddots & \ddots &  \\
  0              & \dots & 0& 1/\sigma_r  \\
\end{pmatrix}
}_{r\times r \text{ nonzero block}}
& \underbrace{\textbf{0}}_{r\times (m-r)} \\
\underbrace{\textbf{0}}_{(n-r)\times r} & \underbrace{\textbf{0}}_{(n-r)\times (m-r)}
\end{bmatrix} U^{\top} \\
& = U \begin{bmatrix}
\underbrace{
\begin{pmatrix}
  1&0   & \cdots & 0 \\
  0       & 1& \ddots & \vdots \\
  \vdots         & \ddots & \ddots &  \\
  0              & \dots & 0& 1  \\
\end{pmatrix}
}_{r\times r \text{ Identity block}}
& \underbrace{\textbf{0}}_{r\times (m-r)} \\
\underbrace{\textbf{0}}_{(m-r)\times r} & \underbrace{\textbf{0}}_{(m-r)\times (m-r)}
\end{bmatrix} U^{\top}\\
& =  \begin{bmatrix}
\underbrace{U_r}_{m\times r} & \underbrace{\textbf{0}}_{(m)\times (m-r)}
\end{bmatrix} U^{\top} \\
& = \sum_{i=1}^r U_{[:,i]} U_{[:,i]}^{\top}.
\]
\ecor

\bprop \label{vectorizedproductsumidentity} Equivalence of writing $(G)^\dag \nabla_\theta L(\theta)$ as a product of summed matrices vs as stacked matrix vector products, i.e.
\[ 
J \rbra*{\frac{1}{d}\sum_{i=1}^d J^{\top}_i \nabla_{f_\theta}^2 \ell_i J_i}^\dag \rbra*{\frac{1}{d}\sum_{i=1}^d J_i^{\top} \nabla_{f_\theta}\ell_i} 
= J\rbra*{J^{\top} \nabla_{f_\theta}^2 \ell(\mathbf{f}_{\theta})J}^\dag J^{\top}\nabla_{f_\theta}\ell(\mathbf{f}_{\theta}).\]
\eprop
\bprf
First, consider the matrix vector product on the right-hand side of \cref{vectorizedproductsumidentity},
\[ \label{e:vectorizedvectorproduct}
\underbrace{J^{\top}}_{p \times dk}\underbrace{\nabla_{f_\theta}\ell(\mathbf{f}_{\theta})}_{dk \times 1} & = \frac{1}{d} \underbrace{ \begin{bmatrix}
   J_1^{\top},
  \, \dots, \,
  J_d^{\top}
\end{bmatrix}}_{p\times dk}\underbrace{ \begin{bmatrix}
   \nabla_{f_\theta} \ell_1
  \\ \vdots \\
  \nabla_{f_\theta}\ell_d
\end{bmatrix}}_{dk\times 1} \\
& = \frac{1}{d} \sum_{i=1}^d J_i^{\top} \nabla_{f_\theta} \ell_i.\] Now consider the matrix product 
\[ \label{e:vectorizedmatrixproduct}
\underbrace{J^{\top}}_{p \times dk} \underbrace{\nabla_{f_\theta}^2 \ell(\mathbf{f}_{\theta})}_{dk \times dk } \underbrace{J}_{dk \times p } & = \underbrace{ \begin{bmatrix}
   J_1^{\top},
  \, \dots, \,
  J_d^{\top}
\end{bmatrix}}_{p\times dk} \frac{1}{d}
\underbrace{\begin{bmatrix}
\underbrace{\boxed{\nabla^2_{f_\theta} \ell_1}}_{k \times k} & 0 & \cdots & 0 \\[6pt]
0 & \boxed{\nabla^2_{f_\theta} \ell_2} & \ddots & \vdots \\[6pt]
\vdots & \ddots & \ddots & 0 \\[6pt]
0 & \cdots & 0 & \boxed{\nabla^2_{f_\theta} \ell_d} 
\end{bmatrix}}_{dk \times dk} \underbrace{ \begin{bmatrix}
   J_1
  \\ \vdots \\
  J_d
\end{bmatrix}}_{dk\times p} \\
& = \frac{1}{d} \begin{bmatrix}
   J_1^{\top},
  \, \dots, \,
  J_d^{\top}
\end{bmatrix} \begin{bmatrix}
   \underbrace{\nabla^2_{f_\theta} \ell_1 J_1}_{k \times p}
  \\ \vdots \\
  \nabla^2_{f_\theta} \ell_d J_d
\end{bmatrix} \\
& = \frac{1}{d}\sum_{i=1}^d J^{\top}_i \nabla_{f_\theta}^2 \ell_i J_i \\
& = G.
\]
Substituting these into the right-hand side of \cref{vectorizedproductsumidentity}, we get:
\[ J\rbra*{J^{\top} \nabla_{f_\theta}^2 \ell(\mathbf{f}_{\theta})J}^\dag J^{\top}\nabla_{f_\theta}\ell(\mathbf{f}_{\theta})& = J \rbra*{\cref{e:vectorizedmatrixproduct}}^\dag \rbra*{\cref{e:vectorizedvectorproduct}}\\
& = J \rbra*{\frac{1}{d}\sum_{i=1}^d J^{\top}_i \nabla_{f_\theta}^2 \ell_i J_i}^\dag \rbra*{\frac{1}{d}\sum_{i=1}^d J_i^{\top} \nabla_{f_\theta}\ell_i},\] as desired.
\eprf


\paragraph{Carefully writing out the last line in \cref{e:AllFunctionSpaceUpdate}.} Recall that $A = \rbra*{\nabla_{f_\theta}^2 \ell(\mathbf{f}_{\theta})}^{1/2} J$, let $r$ be the rank of $A$, and $U$ be the left unitary matrix of the SVD of $A$. Then by \cref{RightPseudoInverseID}
\[-\rbra*{\nabla_{f_\theta}^2 \ell(\mathbf{f}_{\theta})}^{-1/2}&A A^\dag \lt(\nabla_{f_\theta}^2 \ell(\mathbf{f}_{\theta})\rt)^{-1/2}\nabla_{f_\theta}\ell(\mathbf{f}_{\theta})\\
& =  -\rbra*{\nabla^2_{f_\theta} \ell(\mathbf{f}_{\theta})}^{-1/2}
U \begin{bmatrix}
\underbrace{
\begin{pmatrix}
  1&0   & \cdots & 0 \\
  0       & 1& \ddots & \vdots \\
  \vdots         & \ddots & \ddots &  \\
  0              & \dots & 0& 1  \\
\end{pmatrix}
}_{r\times r \text{ Identity block}}
& \underbrace{\textbf{0}}_{r\times (dk-r)} \\
\underbrace{\textbf{0}}_{(dk-r)\times r} & \underbrace{\textbf{0}}_{(dk-r)\times (dk-r)}
\end{bmatrix} U^{\top}
    \frac{1}{d} \underbrace{ \begin{bmatrix}
   \rbra*{\nabla^2_{f_\theta} \ell_1}^{-1/2}\nabla_{f_\theta} \ell_1
  \\ \vdots \\
  \rbra*{\nabla^2_{f_\theta} \ell_d}^{-1/2}\nabla_{f_\theta}\ell_d
\end{bmatrix}}_{dk\times 1}. \]
So, 
\[U^{\top} \rbra*{\nabla_{f_\theta}^2 \ell(\mathbf{f}_{\theta})}^{1/2} \frac{d \mathbf{f}_\theta}{d\tau}
& = - \begin{bmatrix}
	I_r & 0 \\
	0 & 0 
\end{bmatrix} U^{\top}
    \frac{1}{d} \underbrace{ \begin{bmatrix}
   \rbra*{\nabla^2_{f_\theta} \ell_1}^{-1/2}\nabla_{f_\theta} \ell_1
  \\ \vdots \\
  \rbra*{\nabla^2_{f_\theta} \ell_d}^{-1/2}\nabla_{f_\theta}\ell_d
\end{bmatrix}}_{dk\times 1}.
  \]

\[\lt(\nabla_{f_\theta}^2 \ell(\mathbf{f}_{\theta})\rt)^{-1}\nabla_{f_\theta}\ell(\mathbf{f}_{\theta})& = \frac{1}{d} \underbrace{\begin{bmatrix}
\underbrace{\boxed{\rbra*{\nabla^2_{f_\theta} \ell_1}^{-1}}}_{k \times k} & 0 & \cdots & 0 \\[6pt]
0 & \boxed{\rbra*{\nabla^2_{f_\theta} \ell_2}^{-1}} & \ddots & \vdots \\[6pt]
\vdots & \ddots & \ddots & 0 \\[6pt]
0 & \cdots & 0 & \boxed{\rbra*{\nabla^2_{f_\theta} \ell_d}^{-1}} 
\end{bmatrix}}_{dk \times dk} \frac{1}{d}\underbrace{ \begin{bmatrix}
  \nabla_{f_\theta} \ell_1
  \\ \vdots \\
  \nabla_{f_\theta}\ell_d
\end{bmatrix}}_{dk\times 1} \\
& = \frac{1}{d^2} \underbrace{ \begin{bmatrix}
   \rbra*{\nabla^2_{f_\theta} \ell_1}^{-1}\nabla_{f_\theta} \ell_1
  \\ \vdots \\
  \rbra*{\nabla^2_{f_\theta} \ell_d}^{-1}\nabla_{f_\theta}\ell_d
\end{bmatrix}}_{dk\times 1} \]

\subsection{Reachability}
\label{app:reachability}

Let $J = U\Sigma V^{\top}$ be the singular value decomposition of $J$. First, note that the eigenpairs of $G_J = (J^{\top} J)$ satisfy $G_J U_{[:,i]} = \sigma_i^2 U_{[:,i]}$. So, 
\[
U^{\top}_{[:,j]} G_J U_{[:,i]} &= \sigma_i \delta_{i,j}\\
\lt(J U_{[:,i]})^{\top}\rt) \lt(J U_{[:,j]}\rt) & = d \sigma_i \delta_{i,j}\\
 & = \norm{J U_{[:,i]}}^2_2.
\]
We know that $JJ^\dag$ is the orthogonal projector that takes a vector to the column space of $J$. Writing down the projection explicitly
\[J J^\dag  & = J G_J^\dag J^{\top} \\
 & = J U \Sigma^{-1} U^{\top} J^{\top} \\
 & = \sum_{i} \frac{1}{d \sigma_i} \lt(J U_{[:,i]}\rt) \lt(J U_{[:,i]}\rt)^{\top} \\
 & = \sum_i  \dfrac{\lt(J U_{[:,i]}\rt) \lt(J U_{[:,i]}\rt)^{\top}}{\norm{J U_{[:,i]}}_2^2}.
\]

Using this information, we can compute the norm of a vector in the subspace defined by the image of $J$: 
\[JJ^\dag v &= \sum_i  \dfrac{\lt(J U_{[:,i]}\rt) \lt(J U_{[:,i]}\rt)^{\top} v}{d \sigma_i} \\
 & = \sum_i \dfrac{ \lt(\lt(J U_{[:,i]}\rt)^{\top} v\rt)  }{d \sigma_i} \lt(J U_{[:,i]}\rt).
\]
Let $c_i = \dfrac{ \lt(\lt(J U_{[:,i]}\rt)^{\top} v\rt)  }{d \sigma_i} $ which is just a scalar. Then 
\[\norm{JJ^\dag v}_2^2 &= \lt(JJ^\dag v\rt)^{\top} \lt(JJ^\dag v\rt)\\
& = v^{\top} JJ^\dag v \\
& = v^{\top} \sum_i c_i \lt(J U_{[:,i]}\rt)\\
& = \sum_i c_i v^{\top}\lt(J U_{[:,i]}\rt)\\ 
 & = \sum d\sigma_i c_i^2.\]
 
 We define the reachability of a vector by $\norm{JJ^\dag v}_2^2/\norm{v}_2^2$. This tells us how much of $v$ is in Im$(J)$. It is the cosine of the angle between $v$ and the subspace Im$(J)$. If the ratio is 1, then all of $v$ is in the subspace; if it is 0, then none of $v$ is in the subspace.  
 
A similar computation yields the reachability of the GGN. Here, we need to compute in the weighted norm $\norm*{\frac{1}{d} J^{\top} H_\ell J v}_{H_\ell}$:
\[
\norm*{\frac{1}{d} J(J^{\top} H_\ell J)^\dag J^{\top} v}^2_{H_\ell} & = \sum_i d \sigma_i c_i^2,
\]
where $c_i = U_{[:,i]}^{\top} J^{\top} H_\ell v$.

\section{Proofs}
\label{app:proofs}
\bprop Suppose that $\ell$ is strictly convex in $f_\theta$ and twice differentiable. Let
\[
\mathbf{f}_\theta
:=
\begin{bmatrix}
f_\theta(x_1),
\, \dots, \,
f_\theta(x_d)
\end{bmatrix}^{\top}
\in \mathbb{R}^{dk}, \,
\nabla_{f_\theta} \ell(\mathbf{f}_{\theta})= \frac{1}{d} \begin{bmatrix}
   \nabla_{f_\theta} \ell_1,
  \, \dots, \,
  \nabla_{f_\theta}\ell_d
\end{bmatrix}^{\top} \in \reals^{dk}
\]
be the vectorized gradient, and let $\lt(\nabla_{f_\theta}^2 \ell(\mathbf{f}_{\theta})\rt) \in \reals^{dk\times dk}$ be the Hessian of the loss with respect to $\mathbf{f_\theta}$.
\begin{enumerate}
	\item Gauss–Newton descent corresponds to the Newton update direction in function space, restricted to directions reachable through parameter updates.
	\item Moreover, this direction is the unique minimizer of 
\[
\min_{v \in \mathrm{Im}(\nabla_\theta f_\theta)} \frac{1}{2} 
\left\| v + H_\ell^{-1} \nabla_{f_\theta} \ell(f_\theta) \right\|_{H_\ell}^2.
\]
\end{enumerate}
\eprop
\bprf
 Let $J_i:= \nabla_{\theta} f_\theta(\textbf{x}_i)$ and $J = \underbrace{ \begin{bmatrix}
   J_1,
  \, \dots, \,
  J_d
\end{bmatrix}^{\top}}_{dk\times p}$ denote the vectorized matrix containing each sample Jacobian. Note that the vectorized Hessian
\[
\underbrace{\nabla_{f_\theta}^2 \ell(\mathbf{f}_{\theta})}_{dk \times dk}
=\frac{1}{d}
\begin{bmatrix}
\underbrace{\boxed{\nabla^2_{f_\theta} \ell_1}}_{k \times k} & 0 & \cdots & 0 \\
0 & \boxed{\nabla^2_{f_\theta} \ell_2} & \ddots & \vdots \\
\vdots & \ddots & \ddots & 0 \\
0 & \cdots & 0 & \boxed{\nabla^2_{f_\theta} \ell_d}
\end{bmatrix},
\]
is block diagonal and symmetric positive definite. Let $H_\ell := \nabla_{f_\theta}^2 \ell(\mathbf{f}_{\theta})$ and $A := H_\ell^{1/2} J$. This implies that $J =H_\ell^{-1/2} A$.  With this in place, using \cref{lemmaP}, we have
\[
\frac{d \mathbf{f}_\theta}{d\tau} & = J \frac{d \theta}{d\tau} \\
& = - J  \left(G^\dag \nabla_\theta L(\theta)\right)\\
& = -J \rbra*{\frac{1}{d}\sum_{i=1}^d J^{\top}_i H_\ell J_i}^\dag \rbra*{\frac{1}{d}\sum_{i=1}^d J_i^{\top} \nabla_{f_\theta}\ell_i} \\
 & = - J\rbra*{J^{\top} H_\ell J}^\dag J^{\top}\nabla_{f_\theta}\ell(\mathbf{f}_{\theta})\\
 & = -H_\ell^{-1/2}A \rbra*{A^{\top}A}^\dag A^{\top} H_\ell^{-1/2}\nabla_{f_\theta}\ell(\mathbf{f}_{\theta})\\
 & = -H_\ell^{-1/2}A A^\dag \lt(H_\ell^{-1/2}\nabla_{f_\theta}\ell(\mathbf{f}_{\theta})\rt).\\  \label{e:AllFunctionSpaceUpdate}
\]
\eprf 
Note that $f_\theta$'s tangent space is the column space of $J$. Here $A A^\dag$ is the orthogonal projector that takes a vector to the column space of $A$. However, Im$(A) =$ Im$( H_\ell^{1/2} J)$, so the reachable directions correspond exactly to those of the model’s tangent space. 

The calculation above shows that Gauss–Newton computes the update in the model's tangent space that is closest, in the weighted norm $\norm{v}^2_{H_\ell}= v^{\top} H_\ell v $, to the function space Newton direction. 
\bprf
Consider 
\[
v^* \;=\; \argmin_{v \in \operatorname{Im}(J)}
\frac{1}{2} \big\|v + H_\ell^{-1}\nabla_{f_\theta}\ell(\mathbf{f}_{\theta})\big\|_{H_\ell}^2.
\]
The minimizer $v^*$ must satisfy 
\[
\langle v^* + H_\ell^{-1}\nabla_{f_\theta}\ell(\mathbf{f}_{\theta}), w\rangle_{H_\ell} = 0 \quad \forall\, w\in \operatorname{Im}(J).
\]
Since $\operatorname{Im}(J)$ is the column space of $J$, this is equivalent to
\[
J^\top H_\ell v^* = - J^\top \nabla_{f_\theta}\ell(\mathbf{f}_{\theta}).
\]
Writing $v^* = Ju$ for some $u$ yields the normal equations
\[
J^\top H_\ell Ju = -J^\top \nabla_{f_\theta}\ell(\mathbf{f}_{\theta}),
\]
this implies that
$u = -(J^\top H_\ell J)^\dag J^\top \nabla_{f_\theta}\ell(\mathbf{f}_{\theta})$. Multiplying by $J$ we get
\[v^* =-J(J^\top H_\ell J)^\dag J^\top \nabla_{f_\theta}\ell(\mathbf{f}_{\theta}),\]
which is the fourth line on the right-hand side of \cref{e:AllFunctionSpaceUpdate}.
\eprf

\bcor of \cref{generalfunctionspaceflow}. 
$G_J$ gives the closest approximation to $\nabla_{f_\theta} \ell(\mathbf{f}_{\theta})$ in $f_{\theta}$'s tangent space.
\bprf
The argument in \cref{e:AllFunctionSpaceUpdate} becomes
\[\frac{d \mathbf{f}_\theta}{d\tau} & = -J \rbra*{J^{\top}J}^\dag J^{\top} \nabla_{f_\theta}\ell(\mathbf{f}_{\theta})\\
&= -J J^\dag \nabla_{f_\theta}\ell(\mathbf{f}_{\theta}).\\ \label{e:GJfuncspace}
 \]Here, $JJ^\dag$ is the orthogonal projector that takes $\nabla_{f_\theta}\ell(\mathbf{f}_{\theta})$ to the column space of $J$. Note that, unlike the GGN, this projection is in the standard Euclidean norm.
 \eprf
\ecor
Let $J= U \Sigma V^{\top}$ express the singular value decomposition of $J$, then $U$ is a basis for the column space of $J$. Expressing \cref{e:GJfuncspace} in this basis we get
\[U^{\top}\frac{d \mathbf{f}_\theta}{d\tau}  = - \begin{bmatrix}
I_{\text{rank}(J)} & 0 \\
0 & 0 	
 \end{bmatrix}
 U^{\top} \nabla_{f_\theta}\ell(\mathbf{f}_{\theta}).\\ 
 \]

\section{Sketching}
\label{app:sketch}

The updates under $G$, $G_J$, and the Hessian matrix are implemented using the sketching methodology of \cite{SNEGD}, which makes these updates tractable. We do, however, introduce an additional technique: \emph{sketch
sufficiency rank gating}, which ensures that the rank doesn't grow in an unbounded manner. 
For any matrix $M$ used to precondition a gradient descent update via $\theta_{k+1} = \theta_k - M^\dag \nabla_\theta L(\theta)$, the rank-$k$ sketch produces an approximate step
  $\delta_k = U_k \Lambda_k^{-1} U_k^\top \nabla_\theta L(\theta)$ of the ideal step $\delta_* = M^{\dag} \nabla_\theta L(\theta)$, where $U_k$ and
  $\Lambda_k$ are the top-$k$ eigenvectors and eigenvalues from the
  randomized eigendecomposition of $M$. We define the \emph{sufficiency} of the sketch
  as
  \begin{equation}\label{eq:sufficiency}
    S_k \;=\; \frac{\delta_k^\top M\, \delta_k}
                    {\delta_*^\top M\, \delta_*}.
  \end{equation}
  
When the eigendecomposition is exact, $U_k^\top M\, U_k = \Lambda_k$
  and the sufficiency reduces to a weighted projection ratio:
  \begin{equation}\label{eq:sufficiency-exact}
    S_k \;=\;
    \frac{\displaystyle\sum_{i=1}^{k} (u_i^\top \nabla_\theta L(\theta))^2 / \lambda_i}
         {\displaystyle\sum_{i=1}^{n} (u_i^\top \nabla_\theta L(\theta))^2 / \lambda_i},
  \end{equation}
  with $0 \le S_k \le 1$. $S_k =1$ exactly when $\delta^*$ is in the sketched-$k$ subspace (see \cref{app:reachability} for a derivation of reachability: sufficiency is the reachability of the ideal direction vector). The rank adaptively changes in the original sketching algorithm. At each iteration, the sketch is used to compute an eigendecomposition, and the next iteration uses for its sketch size the rank estimate from the previous sketch, where the sketch size is the rank computed as all eigenvalues above a tolerance plus a small oversampling parameter. In practice, this means that the rank can grow unbounded. We introduce sufficiency gating to deal with this problem. When $S_k \geq 1$, the rank is not allowed to increase any further \footnotemark.
  \footnotetext{With approximate eigenvectors, $U_k^\top M\, U_k = \Lambda_k$ isn't necessarily exact. Sketching can underestimate small eigenvalues, so that the numerator of \cref{eq:sufficiency} is too large and $S_k > 1$.}


\section{Additional losses}
\label{app:losses}

\begin{figure}[tbh]
\begin{center}
  \includegraphics[width=\columnwidth]{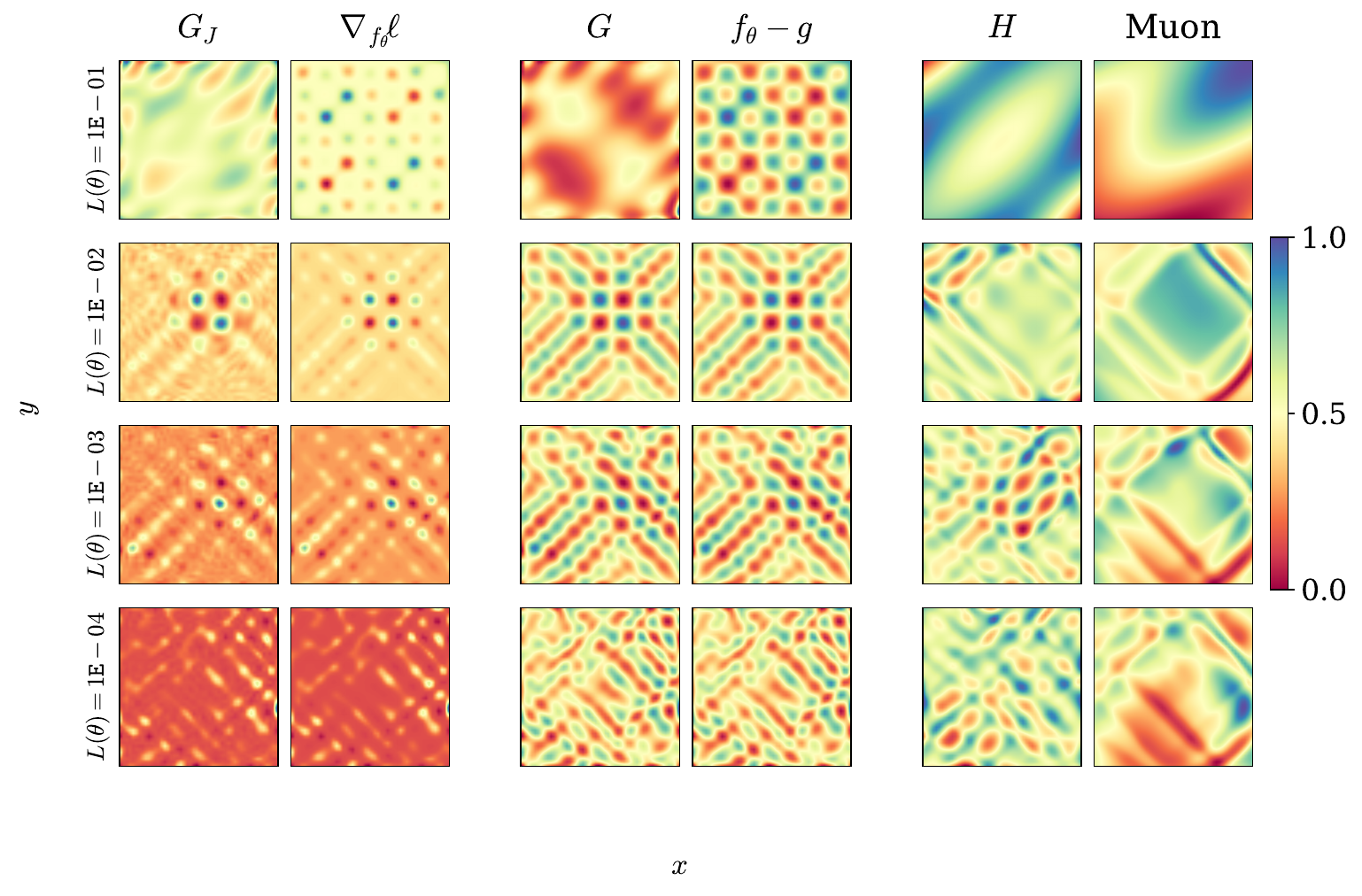}
  \end{center} 
  \caption{Function space update directions for $G_J$, GGN, Newton's method, and Muon compared against the mismatch and the function space gradient of the loss. Each direction is computed using parameters along a Muon optimization path at different loss values. The panels show a heat map over the domain of the mismatch, the function space gradient of the loss, and the direction each optimizer's update points in function space. Directions have been normalized to make visual comparison easier.}
  \label{fig:funcspace_l4}
\end{figure}

\paragraph{A smooth $l^1$ approximation.}
For $\ell = \log\cosh(f_\theta(\mathbf{x}_i)-\mathbf{y}_i)$, we have $\hat\ell(r)=\log\cosh(r)$. In this case the mismatch is still $\psi(\textbf{f}_\theta,\mathbf{y})=\textbf{f}_\theta-\mathbf{y}$, which means $\nabla_{f_\theta} \psi = \textbf{I}$. The mismatch evolves as
\[
G_J: \quad \frac{dr}{d\tau} &=  - \tanh(r) \\
\text{GGN:} \quad \frac{dr}{d\tau} &= - \sinh(r)\cosh(r),
\]
 with solutions
\[
r(\tau) = \text{arc}\sinh\lt(\sinh\lt(r(0)\rt) e^{-\tau}\rt)
\quad\text{and}\quad
r(\tau) = \text{arc}\tanh\lt(\tanh\lt(r(0)\rt) e^{-\tau}\rt),
\]
respectively. Initial behaviour for large $r$ differs, but once $r$ is sufficiently small, both solutions exhibit exponential decay.

\paragraph{Hinge loss.}
For binary labels $\mathbf{y}_i\in\{\pm1\}$, hinge loss corresponds to
$\ell_i=\max\{0,\,1-\mathbf{y}_i f_\theta(\mathbf{x}_i)\}$,
with mismatch
$\psi(f_\theta(\mathbf{x}_i),\mathbf{y}_i)
  =1-\mathbf{y}_i f_\theta(\mathbf{x}_i)$.
Here $\nabla_{f_\theta}\psi(f_\theta(\mathbf{x}_i),\mathbf{y}_i)
  = -\textbf{y}_i$ and
\[
\nabla_{r}\hat{\ell}(r_i)=
\begin{cases}
1, & r_i \geq 0,\\
0, & r_i < 0
\end{cases}.
\] 

The idealized mismatch ODE under $G_J$ is
\[
G_J: \quad \frac{dr_i}{d\tau} = 
\begin{cases}
-1, & r_i \geq 0,\\
\phantom{-}0, & r_i < 0,
\end{cases}
\]
with solution $r_i(\tau) = \max\{r_i(0) - \tau,\, 0\}$. This is uniformly linear decay for all components of the mismatch. The GGN mismatch ODE is not defined, since $H_\ell = 0$ almost everywhere and its inverse does not exist.

  For the hinge loss, under $G_J$, the function space update is $y_i$ for nonzero residuals and zero otherwise. $G_J$ pushes outputs along the line defined by $\textbf{y}$ until  $y_i f_\theta(x_i)=1$. 
   
   \paragraph{LogCosh loss.} We now consider $\ell_i = \log\cosh(f_\theta(\textbf{x}_i)- g(\textbf{x}_i))$. \Cref{sec:error_whitening} tells us that for this loss $G_J$ and $G$ act in very similar ways. For this loss, the function space loss gradient and the mismatch are the same, i.e. $f_\theta- g = \nabla_{f_\theta} \ell $. When the mismatch is sufficiently small $f_\theta- g  \approx \nabla_{f_\theta} \ell $. 

\begin{figure}[hbt]
\begin{center}
  \includegraphics[width=\columnwidth]{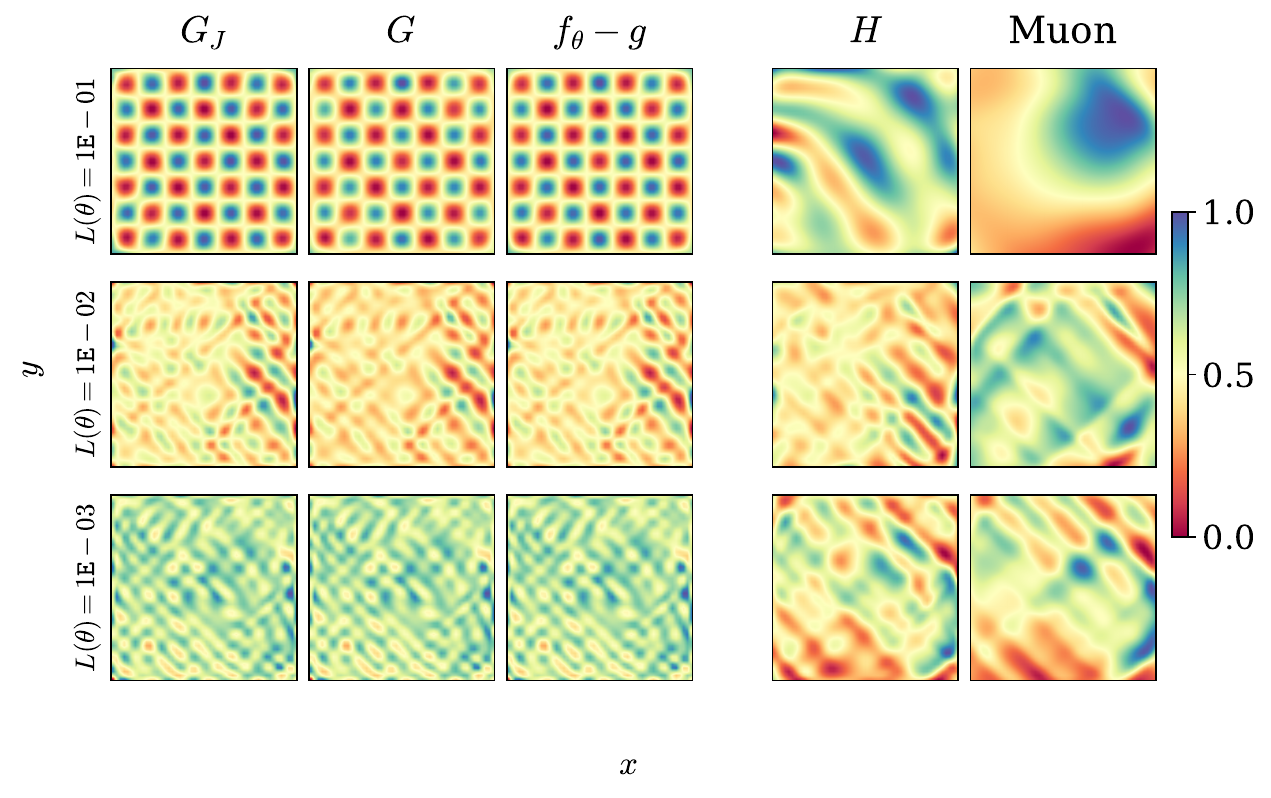}
  \end{center}
  \caption{Function space update directions for $G_J$, GGN, Newton's method, and Muon compared against the mismatch. Each direction is computed using parameters along a Muon optimization path, for loss values near 1E-01 and 1E-04. The neural network is being trained using mean log-cosh loss to approximate the function $g(x,y) = \sin(2 \pi x) \sin(2\pi y) + \sin(7\pi x)\sin(7\pi y)$ on the unit square. The panels show a heat map over the domain of each direction in function space. Directions have been normalized to make visual comparison easier. The cosine similarities are shown in \cref{tab:funcspace-cosine_logcosh}.}\label{fig:funcspace_log_cosh}
\end{figure}

\paragraph{Function space snapshots.} \Cref{fig:funcspace_log_cosh} shows snapshots of function space update directions for all optimizers and the mismatch computed at two loss levels from a Muon training path. The panels for $G_J$ and $G$ are both strikingly similar to $f_\theta- g$. This is confirmed by \cref{tab:funcspace-cosine_logcosh}, which shows that $G_J$, $G$, $\nabla_{f_\theta}\ell $ and $f_\theta -g$ all have cosine similarity 1 for both loss levels. Muon and $H$, on the other hand, align more with each other, but less with the mismatch.

  \begin{table}[tbh]                                                                                            
      \centering                                                                                              
      \caption{Cosine similarities between function space update directions, the mismatch, and the function   
  space loss gradient, computed for parameters along a Muon training                                          
    path, at two loss levels for supervised regression to approximate the function  $\sin(2 \pi x) \sin(2\pi
  y) + \sin(7\pi x)\sin(7\pi y)$ using logcosh loss. The sketch-based directions use a sketch size of    
  18\% of the model's parameters. See \cref{fig:funcspace_log_cosh}.}
      \label{tab:funcspace-cosine_logcosh}                                                                            
      \setlength{\tabcolsep}{3pt}                                                                             
      \begin{subtable}[t]{0.48\textwidth}
      \centering
      \setlength{\tabcolsep}{2pt}
      \begin{tabular}{l|ccccc}
      \multicolumn{6}{c}{\scriptsize$\mathbf{L(\theta) = 1\mathrm{E}{-1}}$} \\
      \toprule
             & \rotatebox{70}{$G_J$} & \rotatebox{70}{$G$} & \rotatebox{70}{$H$} &
  \rotatebox{70}{Muon} &
      \rotatebox{70}{$\nabla_{f_\theta} \ell$} \\
      \midrule
      $G$             &1.00 \\
      $H$                       & .12 & .12 \\
      Muon                      & .06 & .07 & .88 \\
      $\nabla_{f_\theta} \ell$  &1.00 & .99 & .12 & .06 \\
      $f_\theta{-}g$            &1.00 &1.00 & .12 & .07 &1.00 \\
      \bottomrule
      \end{tabular}
      \end{subtable}%
      \hfill
	 \begin{subtable}[t]{0.48\textwidth}
      \centering
      \begin{tabular}{l|ccccc}
      \multicolumn{6}{c}{\scriptsize$\mathbf{L(\theta) = 1\mathrm{E}{-4}}$} \\
      \toprule
             & \rotatebox{70}{$G_J$} & \rotatebox{70}{$G$} & \rotatebox{70}{$H$} &
      \rotatebox{70}{Muon} &
      \rotatebox{70}{$\nabla_{f_\theta} \ell$} \\
      \midrule
      $G$             &1.00 \\
      $H$                       & .70 & .70 \\
      Muon                      & .68 & .68 & .99 \\
      $\nabla_{f_\theta} \ell$  &1.00 &1.00 & .70 & .68 \\
      $f_\theta{-}g$            &1.00 &1.00 & .70 & .68 &1.00 \\
      \bottomrule
      \end{tabular}
      \end{subtable}
    \end{table}
    
\begin{figure}[tbh]
\begin{center}
  \includegraphics[width=0.6\columnwidth]{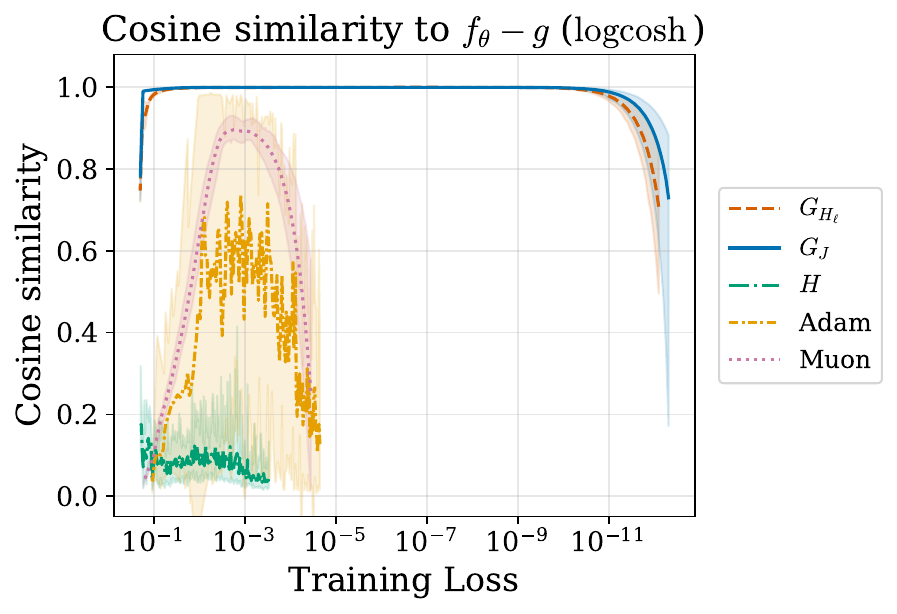}
  \end{center}
\caption{Cosine similarity between each optimizer's function space update direction and the mismatch $\mathbf{f}_\theta - \mathbf{y}$ plotted against training loss for log-cosh loss. Cosine similarities are computed using parameters along each optimizer's respective training path. Solid lines show the mean across 10 random seeds; the shaded regions span the minimum to maximum across seeds.} \label{fig:alignment_log_cosh}
\end{figure}

\paragraph{Training alignment.} \Cref{fig:alignment_log_cosh} shows the cosine similarity to the mismatch as a function of the training loss for all optimizers now along their own training trajectories. The takeaways are very similar to those of quartic loss, except that in this case, both $G$ and $G_J$ align with the mismatch. The Euclidean and $H_\ell$ weighted projections behave nearly identically. Muon's alignment isn't as high as for quartic loss, while Adam's alignment is a bit higher. 

\begin{figure}[tbh]
\begin{center}
  \includegraphics[width=\columnwidth]{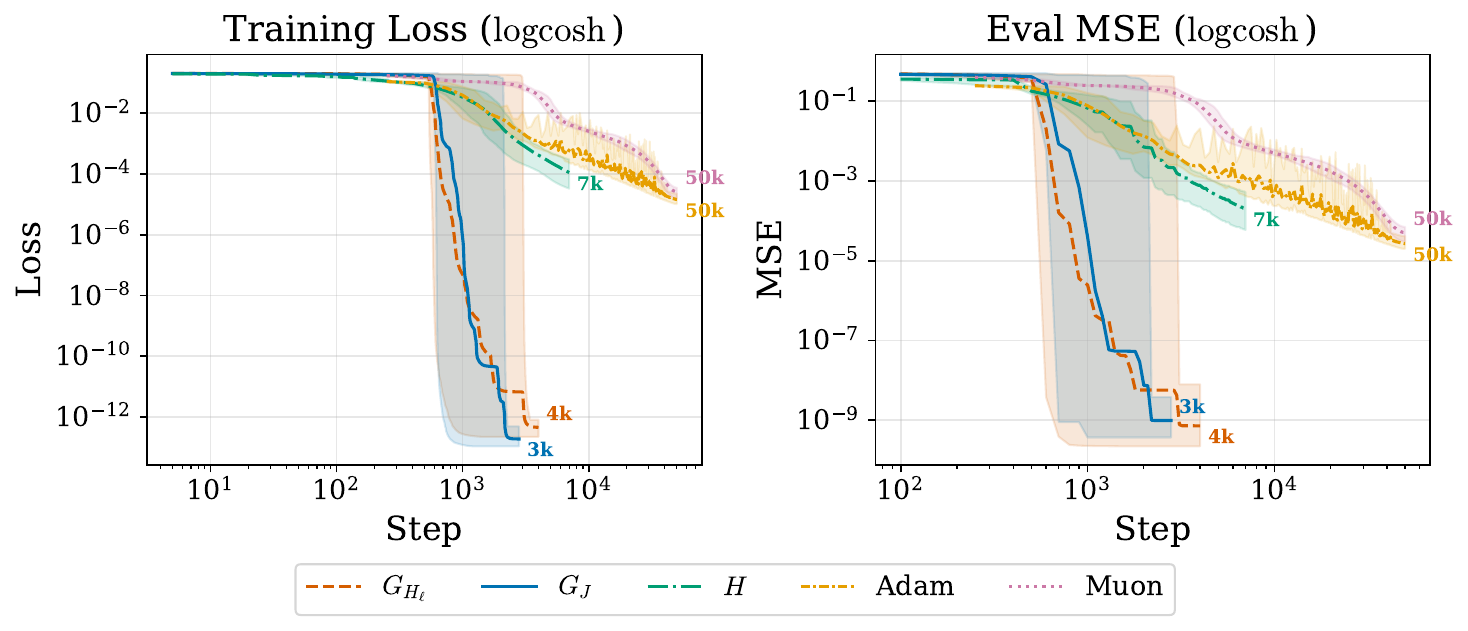}
  \end{center}
 \caption{Convergence plots on a log-log scale showing training loss (left) and evaluation MSE (right) for supervised regression using log-cosh loss $\ell_i = \log\cosh(f_\theta(\mathbf{x}_i) - g(\mathbf{x}_i))$. The neural network is trained to approximate $g(x) = \sin(2\pi x)\sin(2\pi y) + \sin(7\pi x)\sin(7\pi y)$ on the unit square. Labels indicate training iterations. Evaluation MSE is computed on a finer grid of points than the training points. Solid lines show the geometric mean across 10 random seeds; the shaded regions span the minimum to maximum across seeds.} \label{fig:convergence_log_cosh}
\end{figure}

\paragraph{Convergence.} \Cref{fig:convergence_log_cosh} shows the convergence of the loss and the evaluation error for all the optimizers. $G$ and $G_J$ behave very similarly, converging in a small number of steps to errors and losses that are orders of magnitude better than the other optimizers. We report the errors all optimizers achieve in \cref{tab:final-mse}. All optimizers are run across the same 10 random seeds to ensure fair comparison. 
  
\section{Supervised learning case study (continued)}
\label{app:supervised}

\paragraph{Quartic loss.} Define $\ell_i = \frac14 (f_\theta(\textbf{x}_i)- g(\textbf{x}_i))^4$ to be the mean quartic loss $L(\theta)
=
\frac{1}{4d}
\sum_{i=1}^d
\big(f_\theta(\mathbf{x}_i) - g(\mathbf{x}_i)\big)^4
$.

\Cref{tab:funcspace-cosine} shows the pairwise cosine similarities of the function space directions from \cref{fig:funcspace_l4}, with one additional direction shown: the function space update direction under vanilla gradient descent.

\begin{table}[htb!]
    \centering
    \caption{Cosine similarities between function space update directions, the mismatch, and the function space loss gradient, computed for parameters along a Muon training
  path, at two loss levels for supervised regression to approximate the function  $\sin(2 \pi x) \sin(2\pi y) + \sin(7\pi x)\sin(7\pi y)$ using mean quartic loss. The sketch-based directions use a sketch size of 18\% of the model's parameters. See \cref{fig:funcspace_l4}.}
    \label{tab:funcspace-cosine}
    \setlength{\tabcolsep}{3pt}
    \begin{subtable}[t]{0.48\textwidth}
    \centering
    \setlength{\tabcolsep}{2pt}
    \begin{tabular}{l|cccccc}
    \multicolumn{7}{c}{\scriptsize$\mathbf{L(\theta) = 1.2\mathrm{E}{-2}}$} \\
    \toprule
           & \rotatebox{70}{$G_J$} & \rotatebox{70}{$G$} & \rotatebox{70}{$H$} & \rotatebox{70}{Muon} &
    \rotatebox{70}{$\nabla_\theta \ell$} & \rotatebox{70}{$\nabla_{f_\theta} \ell$} \\
    \midrule
    $G$             & .81 \\
    $H$                       & .24 & .31 \\
    Muon                      & .16 & .15 & .60 \\
    $\nabla_\theta \ell$      & .16 & .14 & .60 & .98 \\
    $\nabla_{f_\theta} \ell$  & .92 & .78 & .24 & .16 & .16 \\
    $f_\theta{-}g$            & .81 &1.00 & .31 & .15 & .14 & .78 \\
    \bottomrule
    \end{tabular}
    \end{subtable}%
    \hfill
    \begin{subtable}[t]{0.48\textwidth}
    \centering
    \setlength{\tabcolsep}{2pt}
    \begin{tabular}{l|cccccc}
    \multicolumn{7}{c}{\scriptsize$\mathbf{L(\theta) = 1.1\mathrm{E}{-5}}$} \\
    \toprule
           & \rotatebox{70}{$G_J$} & \rotatebox{70}{$G$} & \rotatebox{70}{$H$} & \rotatebox{70}{Muon} &
    \rotatebox{70}{$\nabla_\theta \ell$} & \rotatebox{70}{$\nabla_{f_\theta} \ell$} \\
    \midrule
    $G$             & .85 \\
    $H$                       & .67 & .90 \\
    Muon                      & .66 & .91 & .98 \\
    $\nabla_\theta \ell$      & .65 & .90 & .97 & .99 \\
    $\nabla_{f_\theta} \ell$  &1.00 & .85 & .67 & .66 & .65 \\
    $f_\theta{-}g$            & .85 &1.00 & .90 & .91 & .90 & .85 \\
    \bottomrule
    \end{tabular}
    \end{subtable}
  \end{table}

We can see the function space behaviour predicted by \cref{sec:error_whitening} play out, $G_J$ has a cosine alignment of 0.92 at loss 1E-02, and 1.0 at loss 1E-05 with $\nabla_{f_\theta} \ell$, while $G$ is in perfect alignment with the mismatch at both levels. The full Hessian points in a markedly different direction than $G$ at loss level 1E-02, with a cosine similarity of only 0.31. This shows that the second-order term $\nabla^2_\theta f_\theta$ dropped by the GGN is nonnegligible and is steering $H$ away from the mismatch. Closer to convergence, $H$ and $G$ align much more closely (0.90) because as residuals shrink, $\nabla^2_\theta f_\theta \to 0$ and the full Hessian converges to the GGN.

 \begin{table}[tbh]
    \centering
    \caption{Final mean squared error (mean $\pm$ std over 10 seeds) for the regression task for each optimizer and loss function, evaluated on the much finer evaluation grid. Table corresponding to \cref{fig:convergence_l4}.}
    \label{tab:final-mse}
    \begin{tabular}{l|cc}
    \toprule
    & Quartic & log-cosh \\
    \midrule
    $G$ (GGN) & $\mathbf{1.7 \pm 0.6 \times 10^{-9}}$  & $1.4 \pm 2.3 \times 10^{-9}$ \\
    $G_J$               & $1.5 \pm 0.4 \times 10^{-4}$  & $\mathbf{1.3 \pm 1.1 \times 10^{-9}}$ \\
    $H$                  & $1.6 \pm 1.0 \times 10^{-4}$  & $2.5 \pm 1.5 \times 10^{-4}$ \\
    Muon                 & $2.9 \pm 1.1 \times 10^{-5}$  & $5.4 \pm 1.2 \times 10^{-5}$ \\
    Adam                 & $9.9 \pm 5.0 \times 10^{-4}$  & $2.9 \pm 0.7 \times 10^{-5}$ \\
    \bottomrule
    \end{tabular}
  \end{table}

    \begin{table}[tbh]                                                                                            
    \centering                                                                                                
    \caption{Hyperparameters for regression experiments.                                             
    All optimizers use the same architecture and initialization.                                              
    Sketch optimizers ($G_J$, $G$, $H$) share identical settings.}                                    
    \label{tab:funcreg-hparams}                                                                               
    \begin{tabular}{@{}ll@{}}                                                                                 
    \toprule                                                                                                  
    \multicolumn{2}{@{}l}{\textbf{Architecture \& data}} \\
    \midrule                                                                                                  
    MLP width / depth & 50 / 6 \quad (12{,}951 params) \\
    Activation / init & Swish / orthogonal, scale $1.8$ \\                                                    
    Train / eval grid & $50^2$ / $150^2$ uniform \\
    Precision & float64\\
    \midrule                                                                                                  
    \multicolumn{2}{@{}l}{\textbf{Sketch optimizers} ($G_J$, $G$, $H$)} \\
    \midrule                                                                                                  
    Rank / oversketch & 75 / 10 \\                                                                            
    Sketch type / tolerance & one-pass / $10^{-14}$ \\                                                        
    Line search & linspace$(0.5,1,6) \cup \{2^{-k}:k\in\text{linspace}(2,30,25)\}$ \\                                     
    Steps & 7{,}001 \\                                                                                        
    \midrule
    \multicolumn{2}{@{}l}{\textbf{Standard optimizers} (Adam, Muon)} \\                                                 
    \midrule                                                                                                  
    LR / schedule / steps & $10^{-3}$ / cosine $\to 0$ / 200{,}001 \\                                         
    $(\beta_1, \beta_2)$ & $(0.9,\, 0.999)$ \\                                                                
    Muon NS steps / $\beta$ & 5 / 0.95 \\                                                                     
    \bottomrule                                                                                               
    \end{tabular}                                                                                                                                                               
  \end{table}

\section{MNIST case study}
\label{app:mnist}

This case study is the classic task of training a neural network to recognize handwritten digits using the MNIST dataset \cite{lecun1998mnist}. 
In \cref{app:supervised}, because the mismatch is a linear function of the model, we have $\nabla_{f_\theta} \psi = \textbf{I}$. In this task, we aim to study what happens when the mismatch dynamics and the function space dynamics are not the same. 
Recall from \cref{sec:error_whitening} that for cross-entropy $\nabla_{f_\theta} \psi = H_{\ell} = F$, the well-known Fisher information matrix.

For this task, we train the network using one-hot labels that encode the correct digit, e.g. a hand-drawn two has a label of $[0,0,1,0,0,0,0,0,0,0]^\top$. The network is trained with cross-entropy loss to predict which of the ten categories a handwritten digit is most likely to fall into. The goal is to compare optimizers, so all hyperparameters are kept the same except the optimizer-specific ones. These are reported in \cref{app:mnist}. The training data consists of 60,000 images, randomly split into batches of size $512$. An epoch is the number of steps required to cycle through all the training data. Accuracy, measured as a percentage, is evaluated on 10,000 images held out from the training data.

\begin{figure}[tbh]
\begin{center}
  \includegraphics[width=0.9\columnwidth]{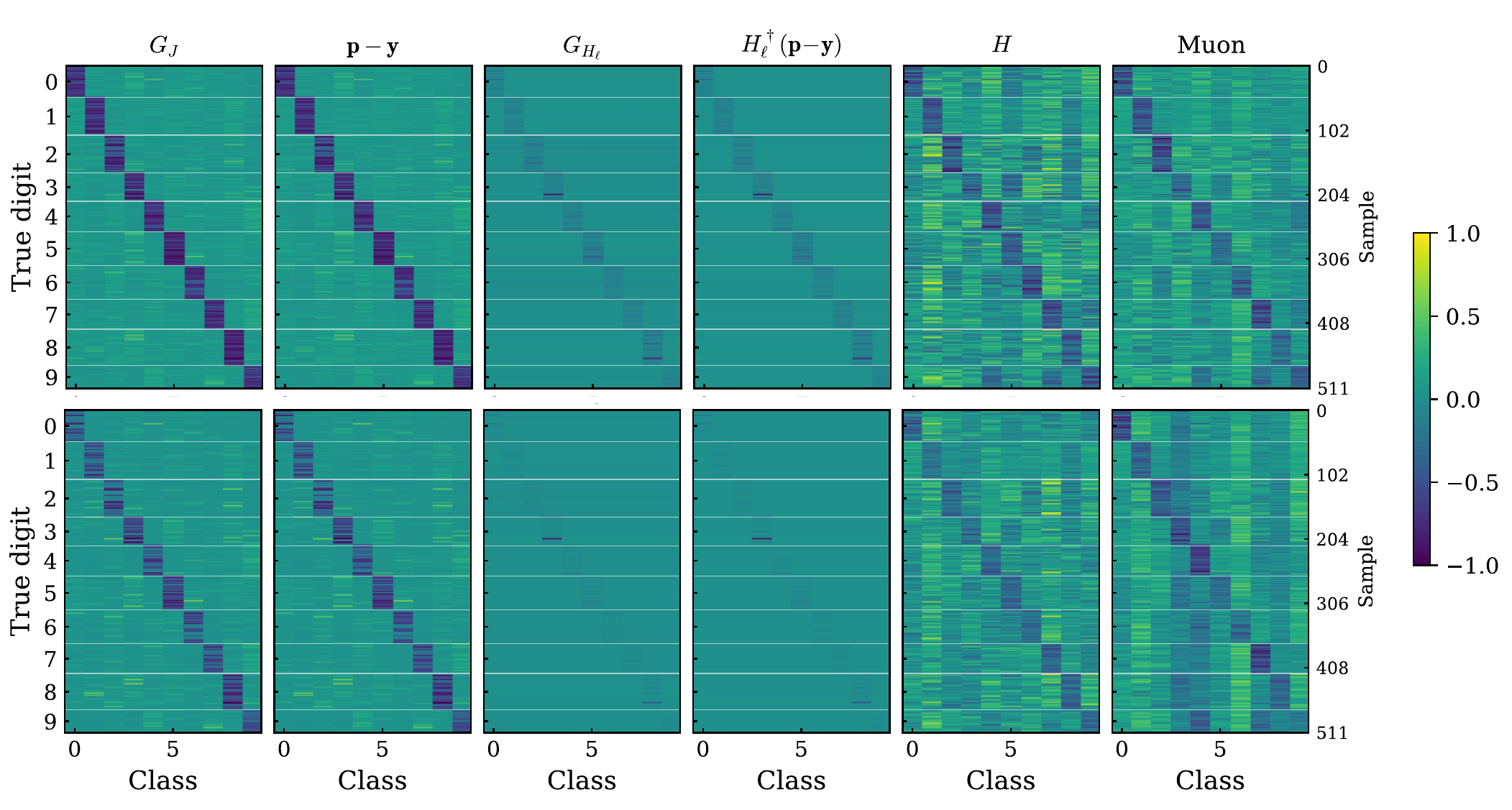}
  \end{center}
 \caption{Function space update directions for $G_J$, $G$, $H$, and Muon, shown alongside the idealized function space update directions for $G_J$ and $G$: the residual $\mathbf{p}-\mathbf{y}$ and $H_\ell^\dag(\mathbf{p}-\mathbf{y})$, respectively. Directions are computed from parameters along a Muon optimization path on MNIST with cross-entropy loss, at accuracies of $75.33\%$ (top) and $84.67\%$ (bottom). Columns correspond to output classes $[0,9]$ and rows are samples from the current training batch, sorted by true label. Colour indicates the signed magnitude of each direction's output for a given sample-class pair, normalized to $[-1,1]$ per panel. Blocks along the diagonal correspond to the correct class for each group of samples, while off-diagonal entries show how the direction affects incorrect classes. Cosine similarities for the panels are shown in \cref{tab:cos_MNIST}.}\label{fig:MNISTHeatmap}
\end{figure}

\paragraph{Function space snapshots.} \Cref{fig:MNISTHeatmap} shows function space updates along a Muon training path for two levels of accuracy on the held-out evaluation data set. $G_J$ and $G$ visually align with their idealized function space update directions. Newton's method and Muon seem to align most closely with each other and exhibit much more off-diagonal activity. These visual patterns are confirmed by the cosine similarities reported in \cref{tab:cos_MNIST} in \cref{app:mnist}. Pseudoinverting $H_\ell$ is numerically unstable, leading to the collapse visible in the heatmaps for $G$ and $H_{\ell}^\dag(\textbf{p}-\textbf{y})$, which lose active entries as accuracy increases.
Theoretically, each of $H_{\ell}$'s $k \times k$ per sample blocks has $k-1$ eigenvalues and a null space made up of the all ones vector. This should lead to a well-defined pseudo-inverse. In practice, $H_{\ell}$'s eigenvalues have a pathological structure and computing the pseudoinverse is not feasible. As the model learns to correctly classify a sample, this process drives the eigenvalues of $H_{\ell}$ arbitrarily close to zero \cite{karakida2021pathological}.

\begin{figure}[hbt]
\begin{center}
  \includegraphics[width=0.5\columnwidth]{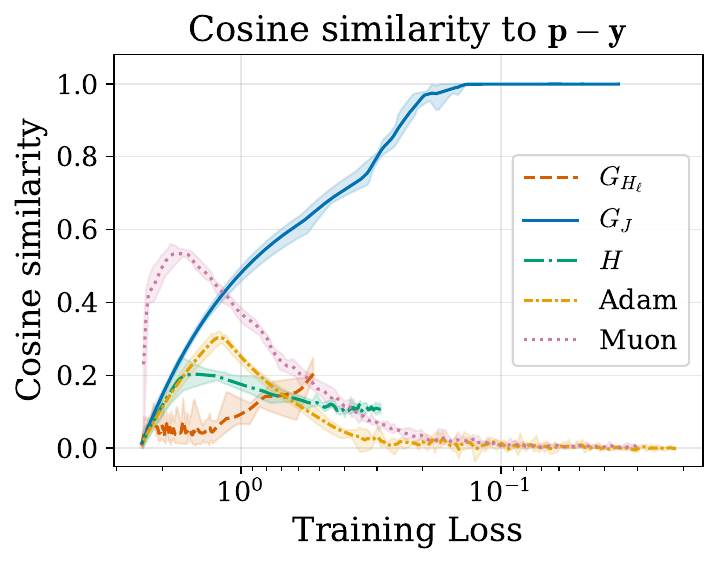}
  \end{center}
 \caption{Cosine similarity between the function space directions across all optimizers and the mismatch as a function of training loss, for training a neural network to predict handwritten digits using the MNIST data set. The solid line shows the mean over five random seeds, while the shaded regions span the min and max across the five seeds.} \label{fig:MNIST_alignment}
\end{figure}

\paragraph{Training alignment.} \Cref{fig:MNIST_alignment} shows the cosine similarity between each optimizer and $\textbf{p}-\textbf{y}$ along its own training trajectory, shown as a function of training loss. Ideally, we would also track each optimizer's alignment with $H_\ell^\dag(\textbf{p}-\textbf{y})$; however, due to the pathological eigenvalue structure of $H_\ell$ discussed above, this couldn't be computed reliably. $G_J$'s cosine similarity with the mismatch builds to perfect alignment and stays there throughout the rest of training. The other optimizers have much lower alignment with the mismatch throughout training. 

\begin{figure}[hbt]
\begin{center}
  \includegraphics[width=\columnwidth]{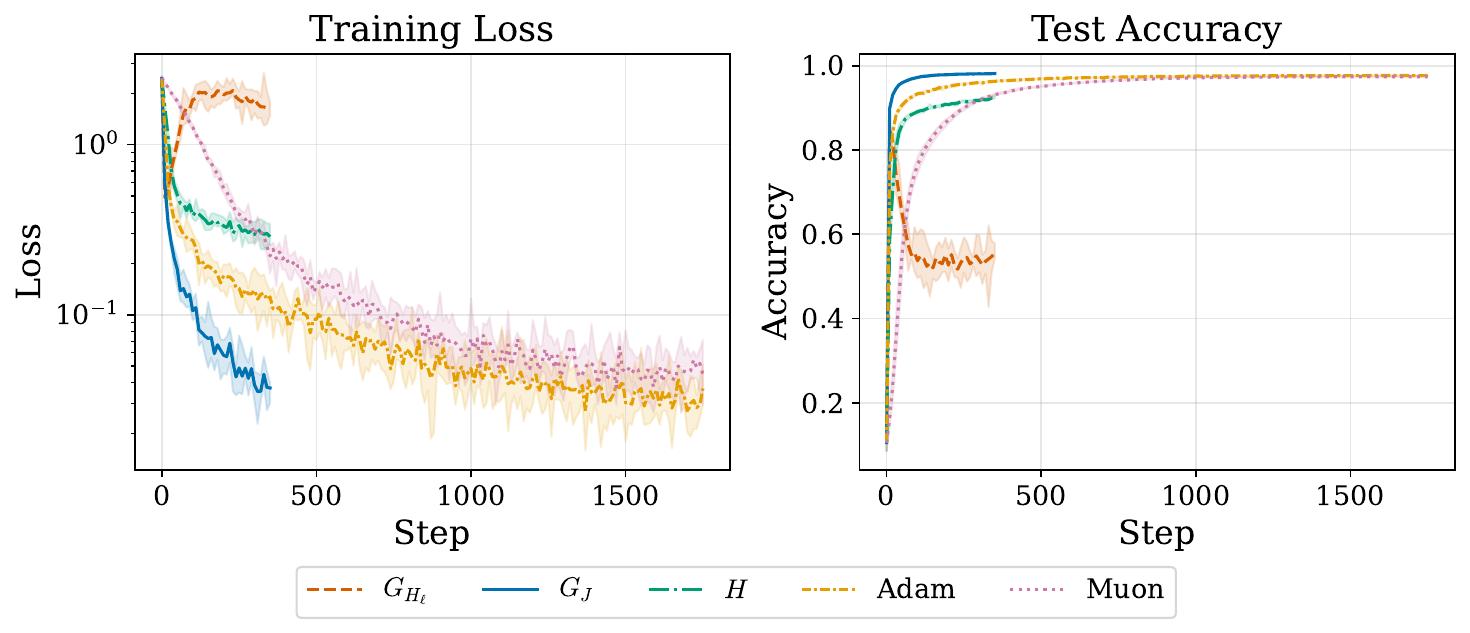}
  \end{center}
 \caption{Convergence plots showing training loss (left) and accuracy on a held back test data set (right) for training with cross-entropy loss to categorize digits from the MNIST dataset. Sketch-based optimizers are trained for three epochs, while Adam and Muon are given fifteen epochs. The solid line shows the geometric mean of the training loss and the standard arithmetic mean of accuracy, averaged over five random seeds. The shaded regions span the min and max across the seeds.} \label{fig:MNIST_convergence}
\end{figure}

\paragraph{Convergence.} \Cref{fig:MNIST_convergence} shows the training loss and test accuracy for all the optimizers. $G_J$'s training loss drops rapidly in the first epoch, followed by a slowing rate of decrease. This is consistent with what the idealized mismatch dynamics suggest. The mismatch decay depends on the eigenvalues of $F$, which become arbitrarily small as training progresses, thereby slowing the decay rate. By contrast, $G$'s mismatch dynamics suggest it should have a similar performance to the fast convergence seen for the regression tasks. Instead, $H_\ell$'s pathological structure leads to training collapse and the lowest final accuracy. Although not visible in \cref{fig:MNIST_convergence}, $G$'s accuracy climbs quickly to around $90\%$ in around 10 training steps and then degrades rapidly. Newton's method performs better than $G$ but only reaches $92.4\%$ accuracy, the second-lowest. Remembering that the Hessian is exactly $G$ with an extra second-order term, $\nabla_\theta^2 f_\theta \nabla_{f_\theta} \ell$, which vanishes near convergence, we might expect Newton to do worse. However, we saw in the last task that Newton's training trajectory seems to favour trajectories that keep this term active. Newton's success over the GGN here suggests the same effect is present, and also that this term may compensate for the ill-conditioning of the Fisher matrix. Adam and Muon are given 15 training epochs, compared to the sketch-based optimizer's 3. Both take more steps to converge but reach high accuracies that are not much below $G_J$'s accuracy.

\begin{table}[htb!]                                                                                                                           
    \centering                                                                                                                                  
    \caption{Cosine similarities between function space update directions for MNIST classification with cross-entropy loss, computed from       
  parameters along a Muon optimization path at two accuracy levels (See \cref{fig:MNISTHeatmap} for heatmaps of these directions). The sketch-based directions use a sketch size of 4.2\% of the model's       
  parameters. Note that: $r = \mathbf{p} - \mathbf{y}$. }
    \label{tab:cos_MNIST}                                                                                                                       
    \setlength{\tabcolsep}{3pt}
    \begin{subtable}[t]{0.48\textwidth}
    \centering
    \setlength{\tabcolsep}{2pt}
    \begin{tabular}{l|ccccc}
    \multicolumn{6}{c}{\scriptsize\textbf{Accuracy = 75.33\%}} \\
    \toprule
           & \rotatebox{70}{$G_J$} & \rotatebox{70}{$r$} & \rotatebox{70}{$G$} & \rotatebox{70}{$H_\ell^\dag r$} &
   \rotatebox{70}{$H$} \\
    \midrule
    $r$                          &1.00 \\
    $G$                           & .76 & .76 \\
    $H_\ell^\dag r$            & .84 & .84 & .91 \\
    $H$                                     & .40 & .40 & .30 & .34 \\
    Muon                                    & .47 & .47 & .32 & .37 & .78 \\
    \bottomrule
    \end{tabular}
    \end{subtable}%
    \hfill
    \begin{subtable}[t]{0.48\textwidth}                                                                                                         
    \centering    
    \begin{tabular}{l|ccccc}
    \multicolumn{6}{c}{\scriptsize\textbf{Accuracy = 84.67\%}} \\
    \toprule                                                                                                                                    
           & \rotatebox{70}{$G_J$} & \rotatebox{70}{$r$} & \rotatebox{70}{$G$} & \rotatebox{70}{$H_\ell^\dag r$} &
   \rotatebox{70}{$H$} \\                                                                                                                       
    \midrule      
    $r$                          &1.00 \\                                                                                              
    $G$                           & .56 & .56 \\
    $H_\ell^\dag r$            & .60 & .61 & .93 \\
    $H$                                     & .33 & .33 & .22 & .24 \\                                                                          
    Muon                                    & .27 & .27 & .17 & .18 & .61 \\
    \bottomrule                                                                                                                                 
    \end{tabular} 
    \end{subtable}                                                                                                                              
  \end{table}

\Cref{tab:mnist-acc} reports the final test accuracies for all the optimizers.
\begin{table}[htb!]
    \centering
    \caption{Best test accuracy (\%, mean $\pm$ std over 5 seeds) for the MNIST classification task using cross-entropy loss. Table corresponding to \cref{fig:MNIST_convergence}.}
    \label{tab:mnist-acc}
    \begin{tabular}{l|c}
    \toprule
    & Accuracy (\%) \\
    \midrule
    $G_J$               & $\mathbf{98.1 \pm 0.1}$ \\
    Adam                 & $97.7 \pm 0.1$ \\
    Muon                 & $97.4 \pm 0.1$ \\
    $H$                  & $92.4 \pm 0.3$ \\
    $G$ (GGN) & $84.9 \pm 1.0$ \\
    \bottomrule
    \end{tabular}
\end{table}

    \begin{table}[tbh]                                                                                                                           
      \centering                                                                                                                                
      \caption{Hyperparameters for MNIST experiments.                                                                            
      All optimizers use the same architecture, initialization, and batch size.                                                                 
      Sketch optimizers ($G_J$, $G$, $H$) share identical settings except where noted.}                                                
      \label{tab:mnist-hparams}                                                                                                                 
      \begin{tabular}{@{}ll@{}}                                                                                                                 
      \toprule                                                                                                                                  
      \multicolumn{2}{@{}l}{\textbf{Architecture \& data}} \\                                                                                   
      \midrule                                                                                                                                  
      MLP width / depth & 128 / 2 \quad (118{,}282 params) \\                                                                                   
      Activation / init & Swish / orthogonal, scale $1.8$ \\                                                                                    
      Loss & Cross-entropy with softmax \\                                                                                                      
      Batch size & 512 \\                                                                                                                       
      Precision & float32  \\                                                                   
      \midrule                                                                                                                                  
      \multicolumn{2}{@{}l}{\textbf{Sketch optimizers} ($G_J$, $G$, $H$)} \\                                                           
      \midrule                                                                                                                                  
      Epochs / steps & 3 / 351 \\                                                                                                               
      Rank & 350 \\                                                                                                                           
      Sketch type & one-pass \\                                                                                                                 
      Tolerance & $10^{-7}$ \\                                                                                                                                                                                           
      Line search & linspace$(0.5,1,6) \cup \{2^{-k}:k\in\text{linspace}(2,30,25)\}$ \\   
      \midrule                                                                                             
      \multicolumn{2}{@{}l}{\textbf{Standard optimizers} (Adam, Muon)} \\
      \midrule                                                                                                                                  
      Epochs / steps & 15 / 1{,}755 \\                                                                                                        
      LR / schedule & $10^{-3}$ / cosine $\to 0$ \\                                                                                             
      Adam $(\beta_1, \beta_2)$ & $(0.9,\, 0.999)$ \\                                                                                           
      Muon NS steps / $\beta$ & 5 / 0.95 \\                                                                                                     
      \bottomrule                                                                                                                               
      \end{tabular}                                                                                                                             
  \end{table}    

\section{PINN case study (continued)}\label{app:PINN}

\subsection{Allen--Cahn equation}                  
  We consider the canonical Allen--Cahn benchmark
  \citep{wang2022respecting,wang2024piratenets, daw2023mitigating} on                                                                                                         
  $\Omega \times [0,T] = [-1, 1] \times [0, 1]$ with                                                                                                   
  diffusion coefficient $\nu = 10^{-4}$:                                                                                                               
  \begin{equation}                                                                                                                                     
      \frac{\partial u}{\partial t}
      + 5\,u^3 - 5\,u                                                                                                                                  
      - \nu\,\frac{\partial^2 u}{\partial x^2}
      \;=\; 0,                                                                                                                                         
      \qquad (x, t) \in \Omega \times (0, T],
  \end{equation}                                                                                                                                       
  with periodic boundary conditions,
  \begin{equation}                                                                                                                                     
      u(x, 0) = x^2 \cos(\pi x), \qquad                                                                                                                
      u(-1, t) = u(1, t) \quad \forall t \in [0, T].                                                                                                   
  \end{equation}                                                                                                                                       
  The reference solution $u^*(x, t)$ is from \cite{wang2023expert} sampled on a                                                                              
  $201 \times 512$ grid in $(t, x)$. We report errors in \cref{tab:PINNerr} relative $\ell_2$ errors
  are $\ell_2$ norm $\lVert u_\theta - u^*\rVert_2 / \lVert u^*\rVert_2$ on                                                                                
  this grid.                                                                                                                                           
                                                                                                                                                       
  \paragraph{PINN composite loss.}                                                                                                                     
  For collocation point sets $\{x_i^{\mathrm{pde}}\}$, $\{x_i^{\mathrm{ic}}\}$,
  $\{x_i^{\mathrm{bc}}\}$ of sizes $N_{\mathrm{pde}}$, $N_{\mathrm{ic}}$,                                                                              
  $N_{\mathrm{bc}}$ and per-component residuals                                                                                                        
  \begin{align}                                                                                                                                       
      r_{\mathrm{pde}}(\theta; x, t)                                                                                                                   
          &= \partial_t u_\theta + 5 u_\theta^3 - 5 u_\theta - \nu\, \partial_x^2 u_\theta,\\                                                          
      r_{\mathrm{ic}}(\theta; x)                                                                                                                       
          &= u_\theta(x, 0) - x^2 \cos(\pi x),\\                                                                                                       
      r_{\mathrm{bc}}(\theta; t)                                                                                                                       
          &= u_\theta(-1, t) - u_\theta(1, t),
  \end{align}                                                                                                                                         
  the pinn Loss is                                                                                               
  \begin{equation}                                                                                                                                     
      L(\theta)                                                                                                                              
      = w_{\mathrm{pde}}\, \tfrac{1}{N_{\mathrm{pde}}}\!\!\sum_i \ell\!\left(r_{\mathrm{pde}}^{(i)}\right)                                             
      + w_{\mathrm{ic}}\, \tfrac{1}{N_{\mathrm{ic}}}\!\!\sum_i \ell\!\left(r_{\mathrm{ic}}^{(i)}\right)                                                
      + w_{\mathrm{bc}}\, \tfrac{1}{N_{\mathrm{bc}}}\!\!\sum_i \ell\!\left(r_{\mathrm{bc}}^{(i)}\right),                                               
  \end{equation}                                                                                                                                       
  with quadrature weights $w_{\mathrm{pde}} = (x_{\max}-x_{\min})(t_{\max}-t_{\min})$,                                                                 
  $w_{\mathrm{ic}} = (x_{\max}-x_{\min})$, $w_{\mathrm{bc}} = (t_{\max}-t_{\min})$                                                                     
  and $\ell(r) = \tfrac{1}{2} r^2$. PDE collocation                                                                                 
  points are resampled at every step via R3                                                                                                            
  adaptive sampling \citep{daw2023mitigating}; IC and BC points are                                                                                    
  held fixed.                                                                                                                                          
                                                                                                                                                       
  \begin{table}[tbh]                                                                                                                                   
  \centering      
  \caption{Hyperparameters for the Allen--Cahn PINN experiments.
  All optimizers use the same architecture, initialization, collocation                                                                                
  schedule, and reference grid. Sketch optimizers ($G$, $H$) share
  identical settings; $G \equiv G_J$ at the $\ell_2$ loss used here.}                                                                                  
  \label{tab:allen-cahn-hparams}                                                                                                                       
  \begin{tabular}{@{}ll@{}}                                                                                                                            
  \toprule                                                                                                                                             
  \multicolumn{2}{@{}l}{\textbf{Architecture \& data}} \\
  \midrule                                                                                                                                             
  MLP width / depth & 20 / 8 \quad (3{,}021 params) \\
  Activation / init & Swish / orthogonal, scale $1.27$ \\                                                                                              
  Loss & mean-square ($\ell_2$) on PDE/IC/BC residuals \\
  Domain & $(x, t) \in [-1, 1] \times [0, 1]$, periodic in $x$ \\                                                                                      
  Train collocation & $30 \times 30$ PDE, 30 IC, 30 BC (R3 resampled) \\                                                                               
  Reference grid & $512 \times 201$  \\                                                                                   
  Precision & float64 \\                                                                                                                               
  \midrule                                                                                                                                             
  \multicolumn{2}{@{}l}{\textbf{Sketch optimizers} ($G$, $H$)} \\
  \midrule                                                                                                                                             
  Rank / oversketch & 100 / 10 \\                                                                                                                      
  Sketch type / tolerance & one-pass / $10^{-14}$ \\                                                                                                               
  Line search & $\mathrm{linspace}(1, 0.5, 6) \cup \{2^{-k} : k \in \mathrm{linspace}(2, 30, 25)\}$ \\                                                 
  Steps & 4{,}001 (G)  and 8,001 (H)\\                                                                                                                                   
  \midrule                                                                                                                                             
  \multicolumn{2}{@{}l}{\textbf{Standard optimizers} (Adam)} \\                                                                                        
  \midrule                                                                                                                                             
  LR / schedule / steps & $10^{-3}$ / cosine $\to 10^{-6}$ / 50{,}001 \\
  $(\beta_1, \beta_2)$ & $(0.9,\, 0.999)$ \\   
  Steps & 200{,}001 \\
  \bottomrule     
  \end{tabular}                                                                                                                                        
  \end{table}  

\begin{table}[tbh]
    \centering
    \caption{Mean $\pm$ std over 5 random seeds for the Allen--Cahn equation.} \label{tab:PINNerr}
    \label{tab:final-metrics}
    \begin{tabular}{l|cc}
    \toprule
         & MSE & Relative \ $\ell_2$ error \\
    \midrule
    $G_J$ & $\mathbf{5.1 \pm 1.3 \times 10^{-9}}$  & $\mathbf{9.97 \pm 1.20 \times 10^{-5}}$ \\
    Adam  & $9.1 \pm 5.1 \times 10^{-2}$            & $4.0 \pm 1.6 \times 10^{-1}$ \\
    $H$   & $4.968 \pm 0.009 \times 10^{-1}$        & $9.929 \pm 0.009 \times 10^{-1}$ \\
    \bottomrule
    \end{tabular}
\end{table}

\section{Reinforcement learning case study (continued)}
\label{app:rl}

\subsection{Double integrator example}

    \begin{table}[tbh]                                                                                            
    \centering                                                                                                
    \caption{Hyperparameters for RL experiment in \cref{sec:rl}                                             
    All optimizers use the same architecture and initialization.                                              
    Sketch optimizers ($G_J$, $G$, $H$) share identical settings. Note we also used other network sizes (64/2, 256/2) and empirically the Gauss-Newton optimizers improved with size, whereas Adam never solved the task and only got worse. Running the non-sketching optimizers for more epochs would also only exacerbate the instability of value estimation, therefore, the final experiments used the same number of iterations as the Gauss-Newton optimizers. We also tried resampling at each epoch (similar to the PINNs example in \cref{sec:pinn}), which improved the estimation error by $G$, but exacerbated the instability of Adam and ultimately had marginal impact on the policy agreement in \cref{fig:value_policy}.}                                    
    \label{tab:rl-hparams}                                                                               
    \begin{tabular}{@{}ll@{}}                                                                                 
    \toprule                                                                                                  
    \multicolumn{2}{@{}l}{\textbf{Architecture \& data}} \\
    \midrule                                                                                                  
    MLP width / depth & 512 / 2  \\
    Activation / init & Swish / orthogonal, scale $1.27$ \\                                                    
    Train / eval grid & 4000 / $121^2$ uniform \\
    Precision & float64\\
    \midrule                                                                                                  
    \multicolumn{2}{@{}l}{\textbf{Sketch optimizers} ($G_J$, $G$, $H$)} \\
    \midrule                                                                                                  
    Rank / oversketch & 96 / 16 \\                                                                            
    Sketch type / tolerance & two-pass / $10^{-5}$ \\                                                        
    Line search & linspace$(0.5,1,6) \cup \{2^{-k}:k\in\text{linspace}(2,30,25)\}$ \\                                     
    Steps & 2000 \\                                                                                        
    \midrule
    \multicolumn{2}{@{}l}{\textbf{Standard optimizers} (Adam, Muon)} \\                                                 
    \midrule                                                                                                  
    LR / schedule / steps & $10^{-3}$ / cosine $\to 0$ / 200 \\                                         
    $(\beta_1, \beta_2)$ & $(0.9,\, 0.999)$ \\                                                                
    Muon NS steps / $\beta$ & 5 / 0.95 \\                                                                     
    \bottomrule                                                                                               
    \end{tabular}                                                                                                                                                                   
  \end{table}

Define the state $s = [ x, \dot{x} ]\T$. The optimal control problem is as follows:
\begin{equation}
    \begin{aligned}
        &\underset{u}{\text{minimize}} && t_f \\
        &\text{subject to} && s(t_0) = x_0 \\
        & && s(t_f) = 0 \\
        & && \Ddot{x}(t) = u(t) \\
        & && \abs{u(t)} \leq 1.
    \end{aligned}
\end{equation}
The solution to this problem is a bang-bang control policy with actions $\{ -1, 1 \}$ everywhere outside the origin.
We turn this into an RL problem by noting the stage cost (reward) can be expressed as:
\begin{align}
    \ell(s) = 
    \begin{cases}
        1 &\text{if } s \neq 0 \\
        0 &\text{otherwise}.
    \end{cases}
\end{align}
For states $s \neq 0$, value targets have the form $\min_{\{-1,1\}} \left\{\ell(s) + V(s')\right\}$ where $s'$ is the transition following $s$ and one of the actions in $\{-1, 1\}$; the target at the origin is $0$.
The policy is therefore derived from one-step planning over the actions $\{-1, 1\}$.
More details on value function learning are given in the following sections.
Hyperparameters for the double integrator experiment are listed in \cref{tab:rl-hparams}.

\begin{figure}[hbt]
\begin{center}
  \includegraphics[width=\columnwidth]{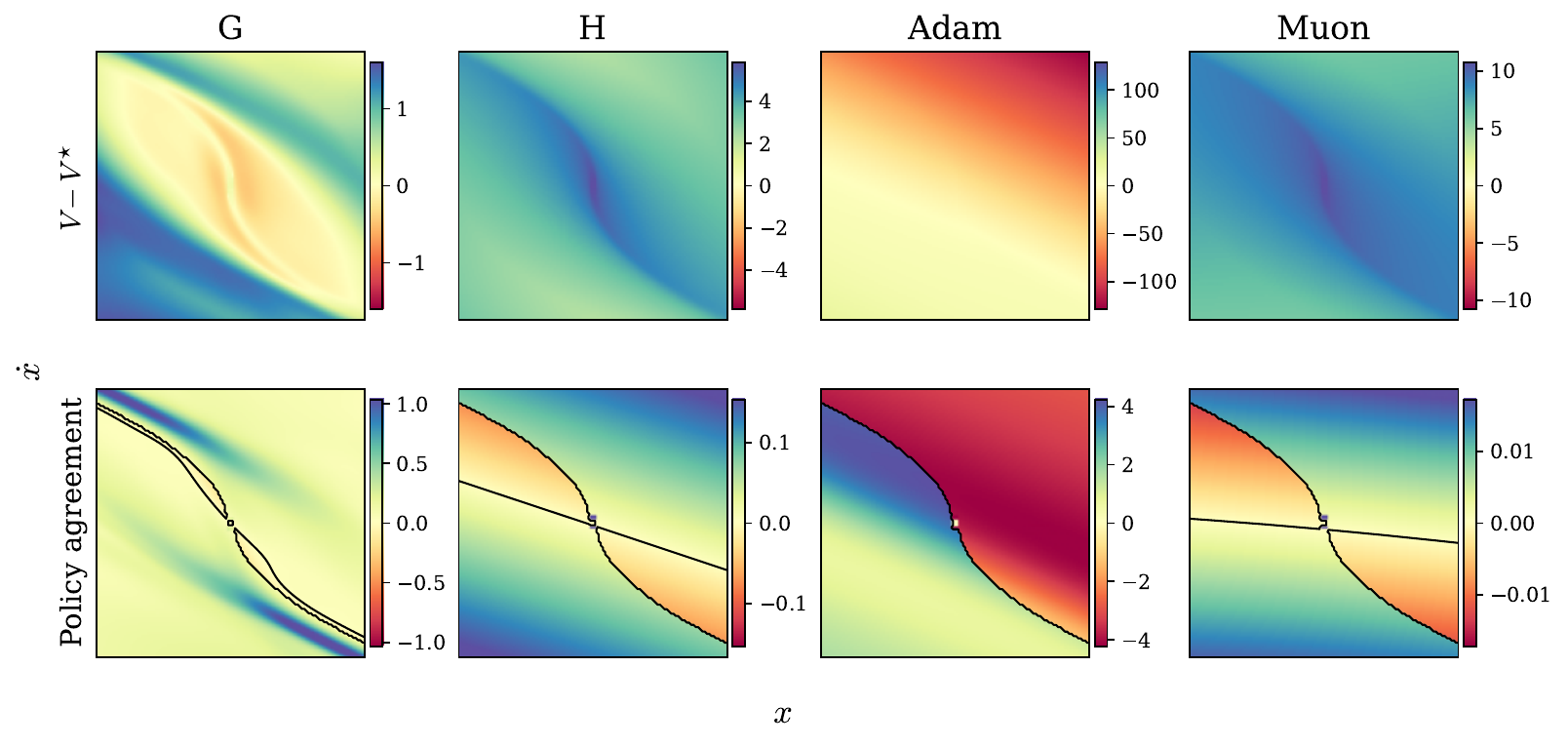}
  \caption{Companion figure to \cref{fig:value_function}.}
\label{fig:value_policy}
\end{center}
\end{figure}

\Cref{fig:value_policy} is a companion to \cref{fig:value_function}.
The top row highlights the error $V_\phi - V^\star$.
Notably, $G$ achieves its lowest error inside the ``switching'' region between control actions; errors outside do not affect the action that is selected as long as the state-action values are correctly ordered.
To quantify this ordering, we introduce the policy agreement in the second row, defined as:
\begin{equation}    
    \left(Q_{\phi}(s,1) - Q_{\phi}(s,-1) \right) \text{sgn}(u^\star),
\end{equation}
where 
\begin{equation}
Q_{\phi}(s,a) = \ell(s) + V_{\phi}(f(s,a)),
\end{equation}
where $f$ is the state transition function and $\text{sgn}(u^\star)$ is the sign of the true optimal action (technically, the sign is redundant for this example in particular).
We see $G$ achieves positive policy agreement nearly uniformly except inside the narrow traced region.
Meanwhile, $H$ and Muon have larger disagreement regions and Adam simply splits the state space along the `S' curve.

\subsection{Crash course on reinforcement learning}

We consider a Markov decision process (MDP) wherein an ``agent'' interacts with an ``environment'' \cite{sutton2018ReinforcementLearning}.
The environment has states $s \in \state$ and actions $a \in \action$ and successive states $s' \in \state$ evolve according to a probability density $s' \sim \pp{p}{s'}{s,a}$.
Additionally, state-action pairs are assigned a ``reward'' based on their utility.
An MDP is an optimization problem over policies:
\begin{equation}
\begin{aligned}
    &\text{maximize} && J(\pi) = \mathbb{E}_{\pi}\left[ \sum_{t=0}^{\infty} \gamma^{t}r (s_t,a_t) \right]\\
    &\text{where} && s_0 \sim p\left(s_0\right)\\
    & && a_t \sim \pp{\pi}{a_t}{s_t}\\
    & &&  s_{t+1} \sim \pp{p}{s_{t+1}}{s_t,a_t}
\end{aligned}
\label{eq:mdpobjective}
\end{equation}
The state-action value function, or $Q$-function, mirrors the objective $J$ but assumes the environment starts at a given $(s, a)$ tuple: $Q^{\pi} (s,a) = \mathbb{E}_{\pi}\left[ \sum_{t=0}^{\infty} \gamma^{t}r (s_t,a_t) \mid s_0 = s, a_0 = a \right]$.
It can be shown that $Q^\pi$ satisfies the following identity:
\begin{equation}
    Q^{\pi} (s,a) = r(s,a) + \gamma \mathbb{E}_{s' \sim \pp{p}{s'}{s,a}, a' \sim \pp{\pi}{a'}{s'}} \left[ Q^{\pi} (s', a') \right].
\label{eq:qfunc}
\end{equation}
\Cref{eq:qfunc} holds for any value function $Q^\pi$; we get the following special case for the optimal value function $Q^\star$:
\begin{equation}
    Q^{\star} (s,a) = r(s,a) + \gamma \mathbb{E}_{s' \sim \pp{p}{s'}{s,a}} \left[ \max_{a' \in \mathcal{A}} Q^{\star} (s', a') \right].
\label{eq:bellman}
\end{equation}
\Cref{eq:bellman} is known as the \emph{Bellman optimality equation}.
It does not make explicit reference to any specific policy as in \cref{eq:qfunc}.
Instead, it implicitly characterizes an optimal policy
\begin{equation}
    \pi^\star (s) = \argmax_{a \in \mathcal{A}} Q^{\star} (s,a),
\label{eq:optpolicy}
\end{equation}
where $\pi^\star$ is not written as a probability density purely for convenience. 
As an aside, in the context of the double integrator problem, we use the state value function, which is completely analogous to the $Q$-function:
\begin{align}
    V^\star (s) &= \max_{a \in  \mathcal{A}} \left\{ r(s,a) + \gamma \mathbb{E}_{s' \sim \pp{p}{s'}{s,a}} \left[ V^\star(s') \right] \right\} \\
    &= \max_{a \in \mathcal{A}} Q(s,a).
\end{align}
Therefore, in the following sections, we may use the state value function in tandem with one-step planning to produce a $Q$-function estimate.

\Cref{eq:bellman} transforms the original infinite-horizon problem of \cref{eq:mdpobjective} into a one-step recursion.
Consequently, if $Q^\star$ is available, then optimal decision-making can be done in a ``greedy'' fashion as in \cref{eq:optpolicy}.
This is because all the planning required towards solving \cref{eq:mdpobjective} is implicitly cached inside $Q^\star$, meaning optimization only over the action variable is required.
In reality, $Q^\star$ is not available; nonetheless \cref{eq:bellman} and \cref{eq:optpolicy} form the basis for many approximate solution methods, discussed next.

\subsection{Approximation landscape}

Finding $Q^\star$ that satisfies \cref{eq:qfunc} is generally intractable for a variety of reasons.
Namely, the dynamics are often unknown, the expectation cannot be evaluated exactly, and the optimization over actions may be expensive \cite{bertsekas1996neuro, bertsekas2022lessons}.
Moreover, \cref{eq:qfunc} must hold for all $(s,a) \in \mathcal{S} \times \mathcal{A}$.
Therefore, these problems are quickly compounded as the state-action dimension increases, a phenomenon known as the ``curse of dimensionality'' \cite{powell2011ApproximateDynamic}.
However, \cref{eq:bellman} and \cref{eq:optpolicy} inspire iterative approximations.

Let $Q_\phi$ be a neural network with parameters $\phi$.
While many algorithms have been developed to tackle \cref{eq:mdpobjective}, a common thread in most cases is the goal of producing $\phi^\star$ such that
\begin{equation}
    Q_{\phi^\star} (s,a) \approx r(s,a) + \gamma \mathbb{E}_{s' \sim \pp{p}{s'}{s,a}} \left[ \max_{a' \in \mathcal{A}} Q_{\phi^\star} (s', a') \right] \quad\forall (s,a) \in \state \times \action.
\label{eq:approxbellman}
\end{equation}
One of the simplest iterative schemes is fitted $Q$-iteration \cite{riedmiller2005NeuralFitted, gordon1999approximate}.
Given a dataset (or batch) of $N$ transition tuples $\mathcal{T} = \{ (s, a, s'), \ldots\}$, the algorithm loops between two steps:
\begin{equation}
\begin{aligned}
    q_i &\leftarrow r(s_i,a_i) + \gamma \max_{a' \in \action}Q_{\phi}(s_i', a')\\
    \phi &\leftarrow \argmin_{\phi} \frac{1}{2N} \sum_{i} \norm{Q_{\phi} (s_i, a_i) - q_i}^2
\end{aligned}
\label{eq:qiter}
\end{equation}
The first line produces ``labels'' for the $Q$-network, which is then optimized in a regression-like way.
This process then repeats.
Intuitively, this process is generally unstable because the targets $q_i$ are bootstrapped estimates of the $Q$-network, so if the maximization is over-optimistic, then this will contaminate the minimizer in the second line.
Errors can then propagate and a meaningful value function approximator will not be obtained.
While fitted $Q$-iteration is generally not a standalone algorithm, this idea is at the core of many algorithms.
To stabilize the learning process, a number of techniques have been introduced, often with significant overlap \cite{hessel2017RainbowCombining, shengyi2022the37implementation}: (clipped) double $Q$-learning \cite{hasselt2010double, fujimoto2018AddressingFunction}, dueling architecture \cite{wang2016dueling}, target networks \cite{lillicrap2015ContinuousControl}, multistep learning \cite{sutton1988LearningPredict, mnih2016AsynchronousMethods}, gradient and objective clipping \cite{schulman2017ProximalPolicy}, trust regions \cite{schulman2015TrustRegion}, (prioritized) experience replay \cite{lin1992reinforcement, mnih2013PlayingAtari, schaul2015prioritized}, delayed updates \cite{fujimoto2018AddressingFunction}, as well as distributional or soft formulations \cite{bellemare2023Distributionalreinforcement, haarnoja2018Softactorcritic}.

\subsection{Experiment on SOAP for approximate dynamic programming}

\begin{figure}[hbt]
\begin{center}
  \includegraphics[width=0.60\columnwidth]{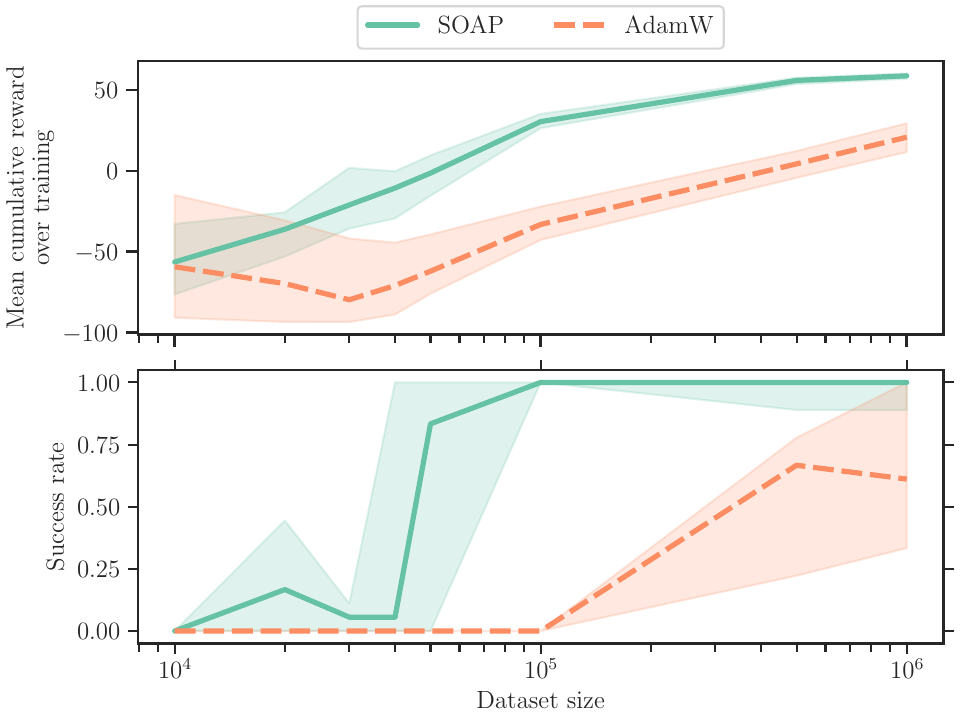}
  \caption{Performance as a function of dataset size. Solid line is the median and shaded regions are the min-max values over $10$ random seeds.}
\label{fig:rl_eval}
\end{center}
\end{figure}

SOAP~\cite{vyas2024soap} is a variant of the Shampoo optimizer~\cite{gupta2018shampoo} that applies Shampoo's updates in a rotated basis. Shampoo can be thought of as a block-diagonal approximation to $G$ \cite{anil2020shampoo,morwani2024shampoo}. We design an experiment based on fitted $Q$-iteration that isolates the effect of this approximate Gauss-Newton optimizer versus AdamW \cite{loshchilov2018decoupled}.
As such, it does not use any of the aforementioned stabilization techniques. Note that for this experiment, the relevant loss is mean-squared error, so $G_J = G$.
To emphasize the regression-like nature of the experiment, the data consists of independent $1$-step tuples $(s, a, s')$, where the $(s,a)$ tuples are uniformly sampled over the state-action space.
That is, there are no rollouts in the environment.
Consequently, the agent must stitch together independent transitions into a global policy at evaluation time.

Our experiment sits comfortably within the deadly triad:
\begin{itemize}
    \item \textbf{Function approximation and bootstrapping.}\quad $Q_\phi$ is represented by a neural network. The same network is used for bootstrapping value estimates. No stabilization techniques are used. Note we do not ever ``freeze'' the parameter $\phi$, meaning every minibatch update step of the form \cref{eq:qiter} changes the target ``labels'' for the next update step.
    \item \textbf{Off-policy targets.}\quad The dataset is static and filled with completely random state-action tuples. Essentially, the entire dataset is exploration data with excellent coverage over the state space. However, fitted $Q$-iteration directly targets the optimal value function, making the problem off-policy. 
\end{itemize}

The environment of interest is the inverted pendulum swing up problem; the reward function is the cosine of the pendulum link's angle.
We apply \cref{eq:qiter} on static datasets and study the effect of the size of the dataset on online evaluation performance.
We do this with the SOAP \cite{vyas2024soap} optimizer as well as AdamW \cite{loshchilov2018decoupled}.
\Cref{fig:rl_eval} shows performance as a function of dataset size.
Each training experiment consists of $200$ epochs over the dataset.
After each epoch, the updated $Q_\phi$-function is evaluated based on how well its greedy policy performs, on average, over $100$ rollouts:
\begin{equation}
    \pi^+ (s) = \argmax_{a \in \mathcal{A}} Q_\phi (s,a).
\label{eq:newpolicy}
\end{equation}
The top figure shows the average cumulative reward over these $200$ epochs.
A higher value indicates faster and more stable training.
The bottom figure translates this performance into a heuristic ``success'' condition: the training experiment is assigned $1$ if the last $10$ evaluation experiments averaged a cumulative reward greater than $30$.

\Cref{fig:rl_eval} shows the SOAP optimizer is able to synthesize a successful policy with $5e4$ data points.
Meanwhile, AdamW was only able to achieve median success over $50\%$ after $5e5$ data points.
Moreover, AdamW was not able to achieve a consistent success rate for larger datasets.
Although this environment is relatively simple, it is striking that SOAP achieves over $10$x data efficiency versus AdamW.
Additionally, this was done in a purely offline, approximate dynamic programming fashion with completely random data.
This demonstrates the possible benefits of improved optimization techniques in a reinforcement learning setting as opposed to heuristic stabilization techniques.
However, more work is required to identify the limits of optimization alone.

\paragraph{Training details.}

Code for this case study is openly available.\footnote{\url{https://github.com/NPLawrence/soapy-rl}}
This experiment, unlike the others, was implemented in PyTorch \cite{paszke2019pytorch} using the NeuroMANCER package \cite{Neuromancer2023} for differentiable programming \cite{drgovna2022differentiable} adapted for neural fitted $Q$-iteration \cite{riedmiller2005NeuralFitted, gordon1999approximate}.
The pendulum has unit length and mass, and friction coefficient of $0.1$.
Although we did not employ stabilization techniques, $Q_\phi$ must be bounded. 
The discrete action space consists of $11$ evenly spaced torque values between $-5$ and $5$.
The architecture had size 256/2 and used GELU activations.
All training experiments were done with a batch size of 512.
We swept over the learning rates $3e-3, 1e-3, 1e-4, 1e-5$.
The final results for SOAP use learning rate $1e-4$; for AdamW we use $1e-5$.

\section{Compute resources and licenses}
\label{app:computers}

The MNIST experiment was performed locally on an Apple M4 Pro laptop with 48 GB of RAM. 
All other experiments, i.e. the function regression and PINN experiments were performed on Google Colab using a single NVIDIA A100-SXM4 GPU with 80 GB of RAM available.
All experiments in \cref{app:rl} were performed locally on an Apple M3 Pro laptop with 18 GB of RAM. No training session took more than an hour on CPU.

This work primarily relied on JAX \cite{jax2018github} (under Apache License 2.0) as well as PyTorch \cite{paszke2019pytorch} (under BSD 3-Clause), NeuroMANCER \cite{Neuromancer2023} (under Battelle Memorial Institute), and SOAP \cite{vyas2024soap} (under MIT).


\end{document}